\documentclass{article} % For LaTeX2e
\usepackage{iclr2021_conference,times}
\usepackage{amsmath,amsfonts,bm}

\def\eqref#1{equation~\ref{#1}}
\def\1{\bm{1}}

\DeclareMathAlphabet{\mathsfit}{\encodingdefault}{\sfdefault}{m}{sl}
\SetMathAlphabet{\mathsfit}{bold}{\encodingdefault}{\sfdefault}{bx}{n}

\usepackage{hyperref}       % hyperlinks
\usepackage{url}            % simple URL typesetting
\usepackage{booktabs}       % professional-quality tables
\usepackage{amsfonts}       % blackboard math symbols
\usepackage{nicefrac}       % compact symbols for 1/2, etc.
\usepackage{microtype}      % microtypography
\usepackage{enumerate}
\usepackage{graphicx}
\usepackage{subfigure}
\usepackage{booktabs}
\usepackage{multirow}
\usepackage{amsmath}
\usepackage{amssymb}
\usepackage{graphicx}
\usepackage{subfigure}
\usepackage{wrapfig}
\usepackage{hyperref}
\usepackage{pifont}
\usepackage{color}
\usepackage[ruled]{algorithm2e}
\usepackage{algorithmic}

\newtheorem{definition}{Definition}[section]

\usepackage{floatrow}
\newfloatcommand{capbtabbox}{table}[][\FBwidth]
\usepackage{tikz}
\newcommand*\circled[1]{\tikz[baseline=(char.base)]{
            \node[shape=circle,draw,inner sep=0.5pt] (char) {\small #1};}}
\newcommand{\term}[1]{\protect\circled{#1}}
\usepackage{array}
\newcolumntype{L}[1]{>{\raggedright\let\newline\\\arraybackslash\hspace{0pt}}m{#1}}
\usepackage{paralist}

\title{Trusted Multi-View Classification}

\author{Zongbo Han, Changqing Zhang, \thanks{ Corresponding author: Changqing Zhang} \\
College of Intelligence and Computing\\
Tianjin University\\
Tianjin, China \\
\texttt{\{zongbo,zhangchangqing\}@tju.edu.cn} \\
\And
Huazhu Fu \\
Inception Institute of Artificial Intelligence \\
Abu Dhabi, UAE \\
\texttt{hzfu@ieee.org} \\
\AND
Joey Tianyi Zhou \\
Institute of High Performance Computing \\
A*STAR, Singapore \\
\texttt{joey.tianyi.zhou@gmail.com}
}

\iclrfinalcopy % Uncomment for camera-ready version, but NOT for submission.
\begin{document}
\maketitle
\begin{abstract}
Multi-view classification (MVC) generally focuses on improving classification accuracy by using information from different views, typically integrating them into a unified comprehensive representation for downstream tasks. However, it is also crucial to dynamically assess the quality of a view for different samples in order to provide reliable uncertainty estimations, which indicate whether predictions can be trusted. To this end, we propose a novel multi-view classification method, termed trusted multi-view classification, which provides a new paradigm for multi-view learning by dynamically integrating different views at an evidence level. The algorithm jointly utilizes multiple views to promote both classification reliability and robustness by integrating evidence from each view. To achieve this, the Dirichlet distribution is used to model the distribution of the class probabilities, parameterized with evidence from different views and integrated with the Dempster-Shafer theory. The unified learning framework induces accurate  uncertainty and accordingly endows the model with both reliability and robustness for out-of-distribution samples. Extensive experimental results validate the effectiveness of the proposed model in accuracy, reliability and robustness.
\end{abstract}

\section{Introduction}
Multi-view data, typically associated with multiple modalities or multiple types of features, often exists in real-world scenarios. State-of-the-art multi-view learning methods achieve tremendous success across a wide range of real-world applications. However, this success typically relies on complex models \citep{wang2015deep,tian2019contrastive, bachman2019learning, zhang2019cpm,hassani2020contrastive}, which tend to integrate multi-view information with deep neural networks. Although these models can provide accurate classification results, they are usually vulnerable to yield unreliable predictions, particularly when presented with views that are not well-represented (\emph{e.g.}, information from abnormal sensors). Consequently, their deployment in safety-critical applications (\emph{e.g.}, computer-aided diagnosis or autonomous driving) is limited. This has inspired us to introduce a new paradigm for multi-view classification to produce trusted decisions.

For multi-view learning, traditional algorithms generally assume an equal value for different views or assign/learn a fixed weight for each view. The underlying assumption is that the qualities or importance of these views are basically stable for all samples. In practice, the quality of a view often varies for different samples which the designed models should be aware of for adaption. For example, in multi-modal medical diagnosis \citep{perrin2009multimodal,sui2018multimodal}, a magnetic resonance (MR) image may be sufficient for one subject, while a positron emission tomography (PET) image may be required for another. Therefore, the decision should be well explained according to multi-view inputs. Typically, we not only need to know the classification result, but should also be able to answer ``How confident is the decision?" and ``Why is the confidence so high/low for the decision?". To this end, the model should provide in accurate uncertainty for the prediction of each sample, and even individual view of each sample.

Uncertainty-based algorithms can be roughly divided into two main categories, \emph{i.e.}, Bayesian and non-Bayesian approaches. Traditional Bayesian approaches estimate uncertainty by inferring a posterior distribution over the parameters \citep{mackay1992bayesian, bernardo2009bayesian, neal2012bayesian}. A variety of Bayesian methods have been developed, including Laplace approximation \citep{mackay1992practical}, Markov Chain Monte Carlo (MCMC) \citep{neal2012bayesian} and variational techniques \citep{graves2011practical, ranganath2014black, blundell2015weight}. However, compared with ordinary neural networks, due to the doubling of model parameters and difficulty in convergence, these methods are computationally expensive. Recent algorithm \citep{gal2016dropout} estimates the uncertainty by introducing dropout \citep{srivastava2014dropout} in the testing phase, thereby reducing the computational cost.  Several non-Bayesian algorithms have been proposed, including
deep ensemble \citep{lakshminarayanan2017simple}, evidential deep learning \citep{sensoy2018evidential} and deterministic uncertainty estimate \citep{van2020uncertainty}. Unfortunately, all of these methods focus on estimating the uncertainty on single-view data, despite the fact that fusing multiple views through uncertainty can improve performance and reliability.

\begin{figure}[!t]
\centering
\subfigure[Overview of the trusted multi-view classification]{
\begin{minipage}[t]{0.675\linewidth}
\centering
\includegraphics[width=1\linewidth,height=0.4\linewidth]{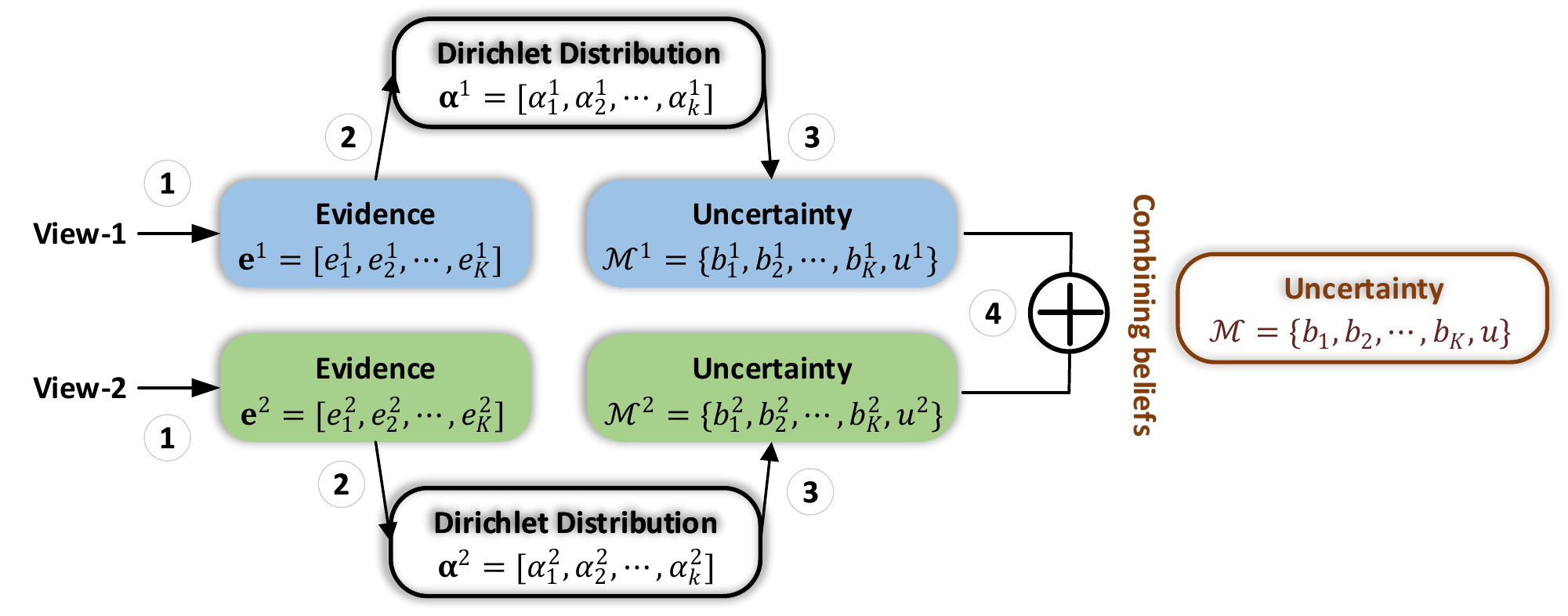}
\centering
\end{minipage}
\label{fig:framework1}}
\subfigure[Combining beliefs]{
\begin{minipage}[t]{0.275\linewidth}
\centering
\includegraphics[width=1\linewidth,height=1\linewidth]{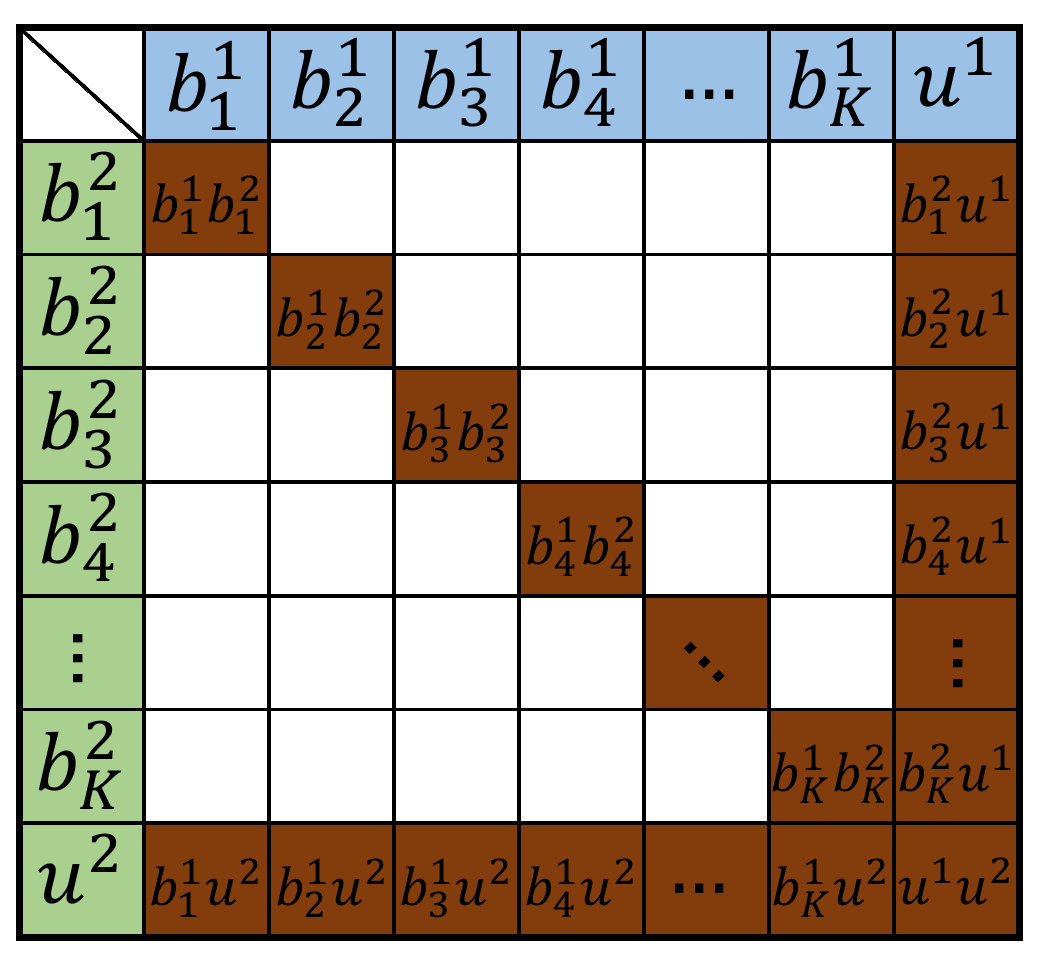}
\end{minipage}
\label{fig:framework2}}
\label{fig:framework}
\caption{Illustration of our algorithm. (a) The evidence of each view is obtained using neural networks (\term{1}). The obtained evidence parameterizes the Dirichlet distribution (\term{2}) to induce the classification probability and uncertainty (\term{3}). The overall uncertainty and classification probability are inferred by combining the beliefs of multiple views based on the DST (\term{4}). The combination rule and an example are shown in Definition~\ref{eq:fusion1} and (b), respectively. Given two sets of beliefs (blue and green blocks), we recombine the compatible parts of the two sets (brown blocks) and ignore the mutually exclusive parts (white blocks) of the two sets to obtain the combined beliefs.}
\end{figure}

In this paper, we propose a new multi-view classification algorithm aiming to elegantly integrate multi-view information for trusted decision making (shown in Fig.~\ref{fig:framework1}). Our model combines different views at an evidence level instead of feature or output level as done previously, which produces a stable and reasonable uncertainty estimation and thus promotes both classification reliability and robustness. The Dirichlet distribution is used to model the distribution of the class probabilities, parameterized with evidence from different views and integrated with the Dempster-Shafer theory. In summary, the specific contributions of this paper are:
\begin{itemize}
\item [(1)] We propose a novel multi-view classification model aiming to provide trusted and interpretable (according to the uncertainty of each view) decisions in an effective and efficient way (without any additional computations and neural network changes), which introduces a new paradigm in multi-view classification.
\item [(2)] The proposed model is a unified framework for promising sample-adaptive multi-view integration, which integrates multi-view information at an evidence level with the Dempster-Shafer theory in an optimizable (learnable) way.
\item [(3)] The uncertainty of each view is accurately estimated, enabling our model to improve classification reliability and robustness.
\item [(4)] We conduct extensive experiments which validate the superior accuracy, robustness, and reliability of our model thanks to the promising uncertainty estimation and multi-view integration strategy.
\end{itemize}

\section{Related Work}
\textbf{Uncertainty-based Learning.} Deep neural networks have achieved great success in various tasks. However since most deep models are essentially deterministic functions, the uncertainty of the model cannot be obtained. Bayesian neural networks (BNNs) \citep{denker1991transforming,mackay1992practical,neal2012bayesian} endow deep models with uncertainty by replacing the deterministic weight parameters with distributions. Since BNNs struggle in performing inference and usually come with prohibitive computational costs, a more scalable and practical approach, MC-dropout \citep{gal2016dropout}, was proposed. In this model, the inference is completed by performing dropout sampling from the weight during training and testing. Ensemble based methods \citep{lakshminarayanan2017simple} train and integrate multiple deep networks and also achieve promising performance.
Instead of indirectly modeling uncertainty through network weights, the algorithm \citep{sensoy2018evidential} introduces the subjective logic theory to directly model uncertainty without ensemble or Monte Carlo sampling. Building upon RBF networks, the distance between test samples and prototypes can be used as the agency for deterministic uncertainty \citep{van2020uncertainty}. Benefiting from the learned weights of different tasks with homoscedastic uncertainty learning, \citep{kendall2018multi} achieves impressive performance in multi-task learning.

\textbf{Multi-View Learning.}  Learning on data with multiple views has proven effective in a variety of tasks. CCA-based multi-view models \citep{hotelling1992relations,akaho2006kernel,wang2007variational,andrew2013deep,wang2015deep,wang2016deep} are representative ones that have been widely used in multi-view representation learning. These models essentially seek a common representation by maximizing the correlation between different views. Considering common and exclusive information, hierarchical multi-modal metric learning (HM3L) \citep{zhang2017hierarchical} explicitly learns shared multi-view and view-specific metrics, while $AE^2$-Nets \citep{zhang2019cpm} implicitly learn a complete (view-specific and shared multi-view) representation for classification. Recently, the methods \citep{tian2019contrastive,bachman2019learning,chen2020simple,hassani2020contrastive} based on contrastive learning have also achieved good performance. Due to its effectiveness, multi-view learning has been widely used in various applications \citep{kiela2018efficient,bian2017revisiting,kiela2019supervised,wang2020makes}.

\textbf{Dempster-Shafer Evidence Theory (DST).}  DST, which is a theory on belief functions, was first proposed by Dempster \citep{dempster1967upper} and is a generalization of the Bayesian theory to subjective probabilities \citep{dempster1968generalization}. Later, it was developed into a general framework to model epistemic uncertainty \citep{shafer1976mathematical}. In contrast to Bayesian neural networks, which indirectly obtain uncertainty through multiple stochastic samplings from weight parameters, DST directly models uncertainty. DST allows beliefs from different sources to be combined with various fusion operators to obtain a new belief that considers all available evidence \citep{sentz2002combination, josang2012interpretation}. When faced with beliefs from different sources, Dempster's rule of combination tries to fuse their shared parts, and ignores conflicting beliefs through normalization factors. A more specific implementation will be discussed later.
\section{Trusted Multi-View Classification}
It has been shown that using a softmax output as confidence for predictions often leads to high confidence values, even for erroneous predictions since the largest softmax output is used for the final prediction \citep{moon2020confidence,van2020uncertainty}. Therefore, we introduce an evidence-based uncertainty estimation technique which can provide more accurate uncertainty and allow us to flexibly integrate multiple views for trusted decision making.

\subsection{Uncertainty and the Theory of Evidence}
In this subsection, we elaborate on evidential deep learning to quantify the classification uncertainty for each of multiple views, which simultaneously models the probability of each class and overall uncertainty of the current prediction. In the context of multi-class classification, Subjective logic (SL) \citep{jsang2018subjective} associates the parameters of the Dirichlet distribution (Definition~\ref{def:dirichlet} in the Appendix) with the belief distribution, where the Dirichlet distribution can be considered as the conjugate prior of the categorical distribution \citep{bishop2006pattern}.

Accordingly, we need to determine the concentration parameters, which are closely related to the uncertainty. We elaborate on the Subjective logic \citep{jsang2018subjective}, which defines a theoretical framework for obtaining the probabilities (belief masses) of different classes and overall uncertainty (uncertainty mass) of the multi-classification problem based on the \emph{evidence} collected from data. Note that \emph{evidence} refers to the metrics collected from the input to support the classification \textcolor{black}{(step \term{1} in Fig.~\ref{fig:framework1})} and is closely related to the concentration parameters of Dirichlet distribution. Specifically, for the $K$ classification problems, subjective logic tries to assign a belief mass to each class label and an overall uncertainty mass to the whole frame based on the \emph{evidence}. Accordingly, for the $v^{th}$ view, the $K+1$ mass values are all non-negative and their sum is one:
\begin{equation}
u^v+\sum_{k=1}^{K} b_{k}^{v}=1,
\label{eq:0}
\end{equation}
where $u^v \geq 0$ and $b_{k}^{v} \geq 0$ indicate the overall uncertainty and the probability for the $k^{th}$ class, respectively.

For the $v^{th}$ view, subjective logic connects the \emph{evidence} $\mathbf{e}^v = [e^v_1, \cdots, e^v_K]$ to the parameters of the Dirichlet distribution $\boldsymbol{\alpha}^v=[\alpha^v_1,\cdots,\alpha^v_K]$ (step \term{2} in Fig.~\ref{fig:framework1}). Specifically, the parameter $\alpha^v_k$ of the Dirichlet distribution is induced from $e^v_k$, \emph{i.e.}, $\alpha_k^v=e^v_k+1$.
Then, the belief mass $b_k^v$ and the uncertainty $u^v$ (step \term{3} in Fig.~\ref{fig:framework1}) are computed as
\begin{equation}
b_{k}^v=\frac{e_k^v}{S^v}=\frac{\alpha_{k}^v-1}{S^v}\quad \text {and} \quad u^v = \frac{K}{S^v},
\label{eq:sl}
\end{equation}
where $S^v=\sum_{i=1}^{K}{(e^v_i+1)}=\sum_{i=1}^{K}{\alpha^v_i}$ is the Dirichlet strength. Eq.~\ref{eq:sl} actually describes the phenomenon where the more evidence observed for the $k^{th}$ category, the greater the probability assigned to the $k^{th}$ class. Correspondingly, the less total evidence observed, the greater the total uncertainty. The belief assignment can be considered as a subjective opinion. Given an opinion, the mean of the corresponding Dirichlet distribution $\hat{\mathbf{p}}^v$ for the class probability $\hat{p}_{k}^v$ is computed as $\hat{p}_{k}^v=\frac{\alpha_{k}^v}{S^v}$ \citep{frigyik2010introduction}.

\textbf{Differences from traditional deep-neural-network classifiers.} Firstly, the output of traditional neural network classifiers can be considered as a point on a simplex, while Dirichlet distribution parametrizes the density of each such probability assignment on a simplex. Therefore, with the Dirichlet distribution, SL models the second-order probability and uncertainty of the output. Secondly, the softmax function is widely used in the last layer of traditional neural network classifiers. However, using the softmax output as the confidence often leads to over-confidence. In our model, the introduced SL can avoid this problem by adding overall uncertainty mass. Existing methods \citep{gal2016dropout,lakshminarayanan2017simple} usually require additional computations during inference to output uncertainty. Since the uncertainty is obtained during the inference stage, it is difficult to seamlessly train a model with high accuracy, robustness and reasonable uncertainty in a unified framework. Accordingly, the limitations underlying existing algorithms (\emph{e.g.}, inability to directly obtain uncertainty) also limits their extension to trusted multi-view classification.
%\end{discussion}
\begin{figure}[!t]
\centering
\subfigure[Confident prediction]{
\centering
\begin{minipage}[t]{0.23\linewidth}
\centering
\includegraphics[width=0.9\linewidth,height=0.75\linewidth]{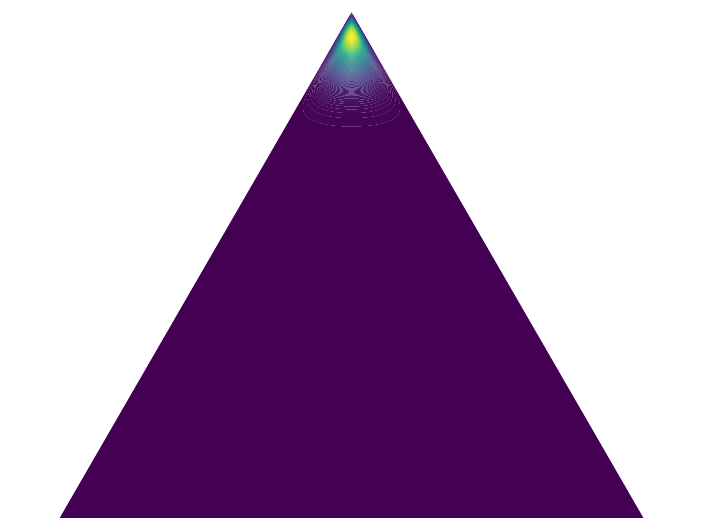}
\centering
\end{minipage}
}
\subfigure[Out of distribution]{
\begin{minipage}[t]{0.23\linewidth}
\centering
\includegraphics[width=0.9\linewidth,height=0.75\linewidth]{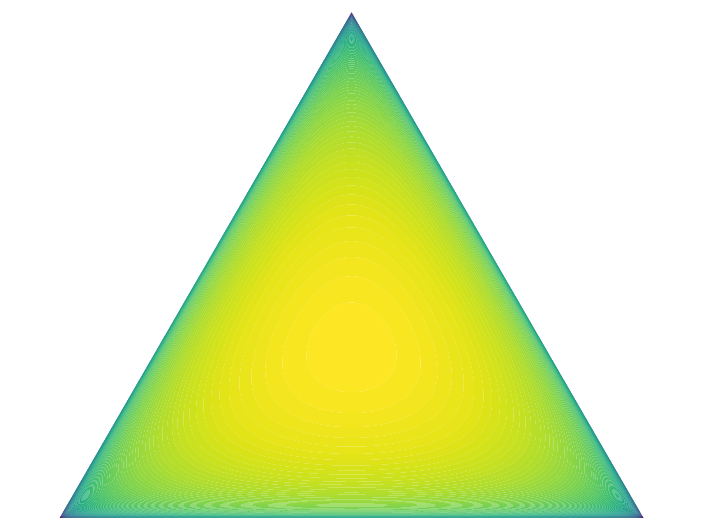}
\end{minipage}}
\centering
\subfigure[High uncertainty]{
\begin{minipage}[t]{0.23\linewidth}
\centering
\includegraphics[width=0.9\linewidth,height=0.75\linewidth]{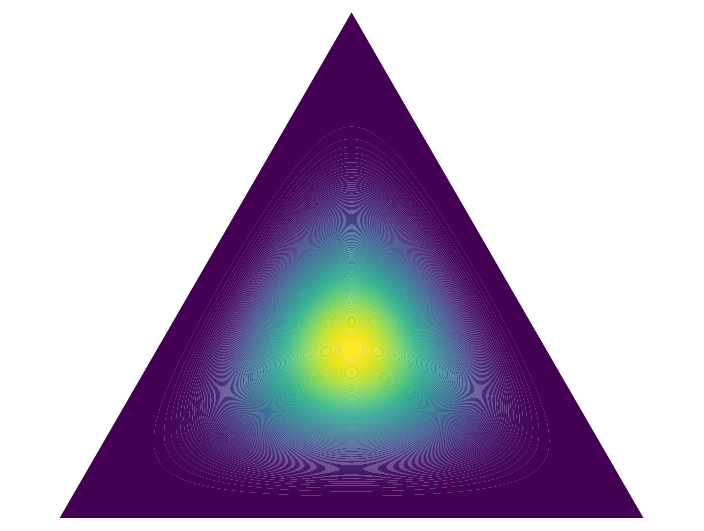}
\end{minipage}}
\subfigure[Opinion $(\mathcal{M})$]{
\centering
\begin{minipage}[t]{0.23\linewidth}
\centering
\includegraphics[width=0.9\linewidth,height=0.8\linewidth]{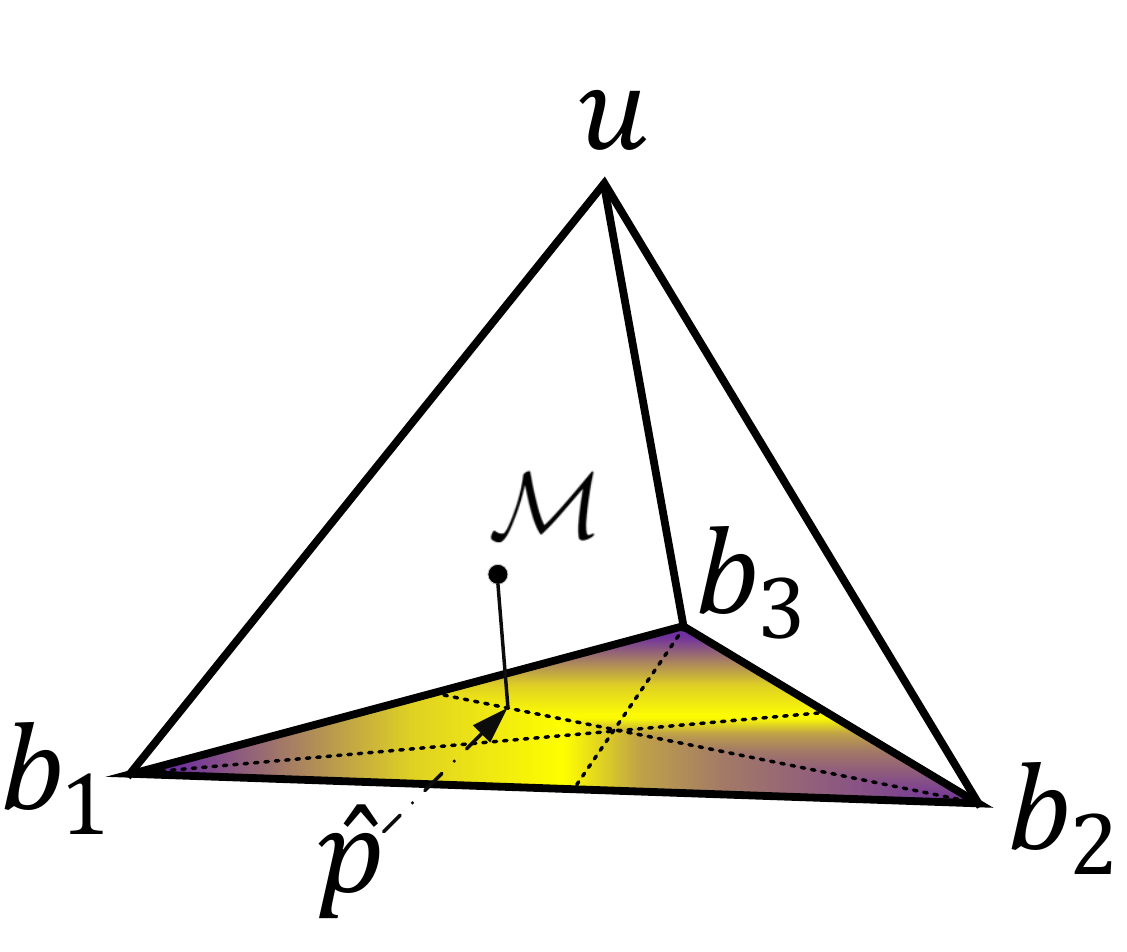}
\centering
\end{minipage}
}
\centering
\caption{Typical examples of Dirichlet distribution. Refer to the text for details.}
\label{fig:dirichlet}
\end{figure}

For clarity, we provide typical examples under a triple classification task to illustrate the above formulation. Let us assume that $\mathbf{e}=\langle 40, 1, 1\rangle$ and accordingly we have $\mathbf{\boldsymbol{\alpha}}=\langle 41, 2, 2\rangle$. The corresponding Dirichlet distribution, shown in Fig.~\ref{fig:dirichlet}(a), yields a sharp distribution centered on the top of the standard 2-simplex. This indicates that sufficient evidence has been observed to ensure accurate classification. In contrast, let us assume that we have the evidence $\mathbf{e}=\langle0.0001, 0.0001, 0.0001\rangle$, which is little evidence for classification. Accordingly, we obtain the Dirichlet distribution parameter $\mathbf{\boldsymbol{\alpha}}=\langle1.0001, 1.0001, 1.0001\rangle$ and the uncertainty mass $u\approx 1$. As shown in Fig.~\ref{fig:dirichlet}(b), in this case, the evidence induces quite a flat distribution over the simplex. Finally, when $\mathbf{e}=\langle 5, 5, 5\rangle$, there is also a high uncertainty, as shown in Fig.~\ref{fig:dirichlet}(c), even though the overall uncertainty is reduced compared to the second case. As shown in Fig.~\ref{fig:dirichlet}(d), we can convert a Dirichlet distribution into a standard 3-simplex (a regular tetrahedron with vertices (1,0,0,0), (0,1,0,0), (0,0,1,0) and (0,0,0,1) in $\mathbf{R}^4$) based on the subjective logic theory (Eq.~\ref{eq:0} and Eq.~\ref{eq:sl}), where the point ($\mathcal{M}$) in the simplex corresponding to $\big\{\{b_k\}_{k=1}^{3}, u\big\}$ indicates an opinion. Accordingly, the expectation value $\hat{\mathbf{p}}$ of the Dirichlet distribution is the projection of $\mathcal{M}$ on the bottom.
\subsection{Dempster's Rule of Combination for Multi-View Classification}
Having introduced evidence and uncertainty for the single-view case, we now focus on their adaptation to data with multiple views. The Dempster–Shafer theory of evidence allows evidence from different sources to be combined arriving at a degree of belief (represented by a mathematical object called the belief function) that takes into account all the available evidence (see Definition~\ref{def2}). Specifically, we need to combine $V$ independent sets of probability mass assignments $\{\mathcal{M}^v\}_1^V$, where $\mathcal{M}^v=\big\{\{b_k^v\}_{k=1}^{K}, u^v\big\}$, to obtain a joint mass $\mathcal{M}=\big\{\{b_k\}_{k=1}^{K}, u\big\}$ (step \term{4} in Fig.~\ref{fig:framework1}).
\begin{definition} (\textbf{Dempster's combination rule for two independent sets of masses})
\label{def2} The combination (called the joint mass) $\mathcal{M}=\big\{\{b_k\}_{k=1}^{K}, u\big\}$ is calculated from the two sets of masses $\mathcal{M}^1= \big\{\{b_k^1\}_{k=1}^{K}, u^1\big\}$ and $\mathcal{M}^2=\big\{\{b_k^2\}_{k=1}^{K}, u^2\big\}$ in the following manner:
\begin{equation}
\mathcal{M}=\mathcal{M}^1\oplus \mathcal{M}^2.
\end{equation}
The more specific calculation rule can be formulated as follows:
\begin{equation}
\begin{aligned}
& b_{k}=\frac{1}{1-C}( b^1_kb^2_k + b^1_ku^2 + b^2_ku^1), u = \frac{1}{1-C}u^1u^2,
\label{eq:fusion1}
\end{aligned}
\end{equation}
where $C = \sum_{i\neq j}{b^1_i b^2_j}$ is a measure of the amount of conflict between the two mass sets (the white blocks in Fig.~\ref{fig:framework2}), and the scale factor $\frac{1}{1-C}$ is used for normalization.
\end{definition}
The joint opinion $\mathcal{M}$ is formed based on the fusion of opinions $\mathcal{M}^1$ and $\mathcal{M}^2$. The joint belief mass of class $k$ ($b_k$) and overall uncertainty ($u$) correspond to the brown blocks in Fig.~\ref{fig:framework2}. Intuitively, the combination rule ensures: (1) when both views are of high uncertainty (large $u^1$ and $u^2$), the final prediction must be of
low confidence (small $b_k$); (2) when both views are of low uncertainty (small $u^1$ and $u^2$), the final prediction may be of high confidence (large $b_k$); (3) when only one view is of low uncertainty (only $u^1$ or $u^2$ is large), the final prediction only depends on the confident view.

Then, given data with $V$ different views, we can obtain the above-mentioned mass for each view. Afterwards, we can combine the beliefs from different views with Dempster's rule of combination. Specifically, we fuse the belief mass and uncertainty mass between different views with the following rule:
\begin{equation}
\begin{aligned}
\mathcal{M}=\mathcal{M}^1\oplus \mathcal{M}^2 \oplus \cdots \mathcal{M}^V.
\label{eq:fusion2}
\end{aligned}
\end{equation}
After obtaining the joint mass $\mathcal{M}=\big\{\{b_k\}_{k=1}^{K}, u\big\}$, according to Eq.~\ref{eq:sl}, the corresponding joint evidence from multiple views and the parameters of the Dirichlet distribution are induced as
\begin{equation}
\begin{aligned}
S = \frac{K}{u} , e_k =  b_k \times S \text{ and } \alpha_k = e_k+1.
\label{eq:sl2}
\end{aligned}
\end{equation}
Based on the above combination rule, we can obtain the estimated multi-view joint evidence $\mathbf{\boldsymbol{e}}$ and the corresponding parameters of joint Dirichlet distribution $\mathbf{\boldsymbol{\alpha}}$ to produce the final probability of each class and the overall uncertainty.

{\textbf{Advantages of using subjective logic compared with softmax.} Compared with softmax output, using subjective uncertainty is more suitable for the fusion of multiple decisions. Subjective logic provides an additional mass function ($u$) that allows the model distinguish between a lack of evidence. In our model, subjective logic provides the degree of overall uncertainty of each view, which is important for trusted classification and interepretability to some extent.}
\subsection{Learning to Form Opinions}
%In this section, we will introduce how to train a neural network to obtain the above $\{M^v\}_{v=1}^V$ and $M$.
In this section, we will discuss how to train neural networks to obtain evidence for each view, which can then be used to obtain the corresponding masses $\{\mathcal{M}^v\}_{v=1}^V$ and $\mathcal{M}$. The neural networks can capture the evidence from input to induce a classification opinion \citep{kiela2018efficient}, and the conventional neural-network-based classifier can be naturally transformed into the evidence-based classifier with minor changes. Specifically, the softmax layer of a conventional neural-network-based classifier is replaced with an activation function layer (\emph{i.e.}, RELU) to ensure that the network outputs non-negative values, which are considered as the evidence vector $\boldsymbol{e}$. Accordingly, the parameters of the Dirichlet distribution can be obtained.

For conventional neural-network-based classifiers, the cross-entropy loss is usually employed:
\begin{equation}
    \mathcal{L}_{c e} = -\sum_{j=1}^{K} y_{i j}\log (p_{i j}),
    \label{eq:crossentropy}
\end{equation}
where $p_{i j}$ is the predicted probability of the $i$th sample for class $j$. For our model, given the evidence of the $i$th sample obtained through the evidence network, we can get the parameter $\boldsymbol{\alpha}_i$ (\emph{i.e.}, $\boldsymbol{\alpha}_i^v=\mathbf{e}^i_i+1$) of the Dirichlet distribution and form the multinomial opinions $D(\mathbf{p}_i|\boldsymbol{\alpha}_i)$, where $\mathbf{p}_i$ is the class assignment probabilities on a simplex.
After a simple modification on Eq.~\ref{eq:crossentropy}, we have the adjusted cross-entropy loss:
\begin{equation}
\mathcal{L}_{ace}(\mathbf{\boldsymbol{\alpha}}_{i})=\int\left[\sum_{j=1}^{K}-y_{i j} \log \left(p_{i j}\right)\right] \frac{1}{B\left(\boldsymbol{\alpha}_{i}\right)} \prod_{j=1}^{K} p_{i j}^{\alpha_{i j}-1} d \mathbf{p}_{i}=\sum_{j=1}^{K} y_{i j}\left(\psi\left(S_{i}\right)-\psi\left(\alpha_{i j}\right)\right),
\label{eq:ace}
\end{equation}
where $\psi(\cdot)$ is the digamma function. 
Eq.~\ref{eq:ace} is the integral of the cross-entropy loss function on the simplex determined by $\mathbf{\boldsymbol{\alpha}}_i$. The above loss function ensures that the correct label of each sample generates more evidence than other classes, however, it cannot guarantee that less evidence will be generated for incorrect labels. That is to say, in our model, we expect the evidence for incorrect labels to  shrink to 0. To this end, the following KL divergence term is introduced:
\begin{equation}
\begin{array}{l}
K L\left[D\left(\mathbf{p}_{i} | \tilde{\boldsymbol{\alpha}}_{i}\right) \| D\left(\mathbf{p}_{i} | \mathbf{1}\right)\right] \\
\quad=\log \left(\frac{\Gamma\left(\sum_{k=1}^{K} \tilde{\alpha}_{i k}\right)}{\Gamma(K) \prod_{k=1}^{K} \Gamma\left(\tilde{\alpha}_{i k}\right)}\right)+\sum_{k=1}^{K}\left(\tilde{\alpha}_{i k}-1\right)\left[\psi\left(\tilde{\alpha}_{i k}\right)-\psi\left(\sum_{j=1}^{K} \tilde{\alpha}_{i j}\right)\right],
\end{array}
\end{equation}
where $\tilde{\boldsymbol{\alpha}}_i = \mathbf{y}_i+(1-\mathbf{y}_i)\odot\mathbf{\boldsymbol{\alpha}}_i$ is the adjusted parameter of the Dirichlet distribution which can avoid penalizing the evidence of the groundtruth class to 0, and $\Gamma(\cdot)$ is the \emph{gamma} function.

Therefore, given parameter $\boldsymbol{\alpha}_i$ of the Dirichlet distribution for each sample $i$, the sample-specific loss is
\begin{equation}
\mathcal{L}(\mathbf{\boldsymbol{\alpha}}_i)= \mathcal{L}_{ace}(\mathbf{\boldsymbol{\alpha}}_i)+\lambda_{t} K L\left[D\left(\mathbf{p}_{i} | \tilde{\boldsymbol{\alpha}}_{\boldsymbol{i}}\right) \| D\left(\mathbf{p}_{i} |\mathbf{1}\right)\right],
\label{eq:la}
\end{equation}
where $\lambda_{t}>0$ is the balance factor. In practice, we can gradually increase the value of $\lambda_{t}$ so as to prevent the network from paying too much attention to the KL divergence in the initial stage of training, which may result in a lack of good exploration of the parameter space and cause the network to output a flat uniform distribution.

To ensure that all views can simultaneously form reasonable opinions and thus improve the overall opinion, we use a multi-task strategy with following overall loss function:
\begin{equation}
\begin{aligned}
\mathcal{L}_{overall} = \sum_{i=1}^{N}\left[\mathcal{L}(\mathbf{\boldsymbol{\alpha}}_i) + \sum_{v=1}^{V}\mathcal{L}(\mathbf{\boldsymbol{\alpha}}_i^v)\right].
\label{eq:loss}
\end{aligned}
\end{equation}
The optimization process for the proposed model is summarized in Algorithm~\ref{alg:alg1} (in the Appendix).
%-------------------------------------------------
\section{Experiments}
\subsection{Experimental Setup}
In this section, we conduct experiments on six real-world datasets: Handwritten\footnote{https://archive.ics.uci.edu/ml/datasets/Multiple+Features}, CUB \citep{wah2011caltech}, Caltech101 \citep{fei2004learning}, PIE\footnote{http://www.cs.cmu.edu/afs/cs/project/PIE/MultiPie/Multi-Pie/Home.html}, Scene15 \citep{fei2005bayesian} and HMDB \citep{kuehne2011hmdb}. We first compare our algorithm with single-view classifiers to validate the effectiveness of our algorithm in utilizing multiple views. Then, we apply existing classifiers to multi-view features and conduct experiments under different levels of noise to investigate their ability in identifying multi-view OOD samples. Details of these datasets and experimental setting can be found in the appendix.

\begin{wrapfigure}{!tr}{0.35\textwidth}
\centering
\includegraphics[width=1\linewidth,height=0.75\linewidth]{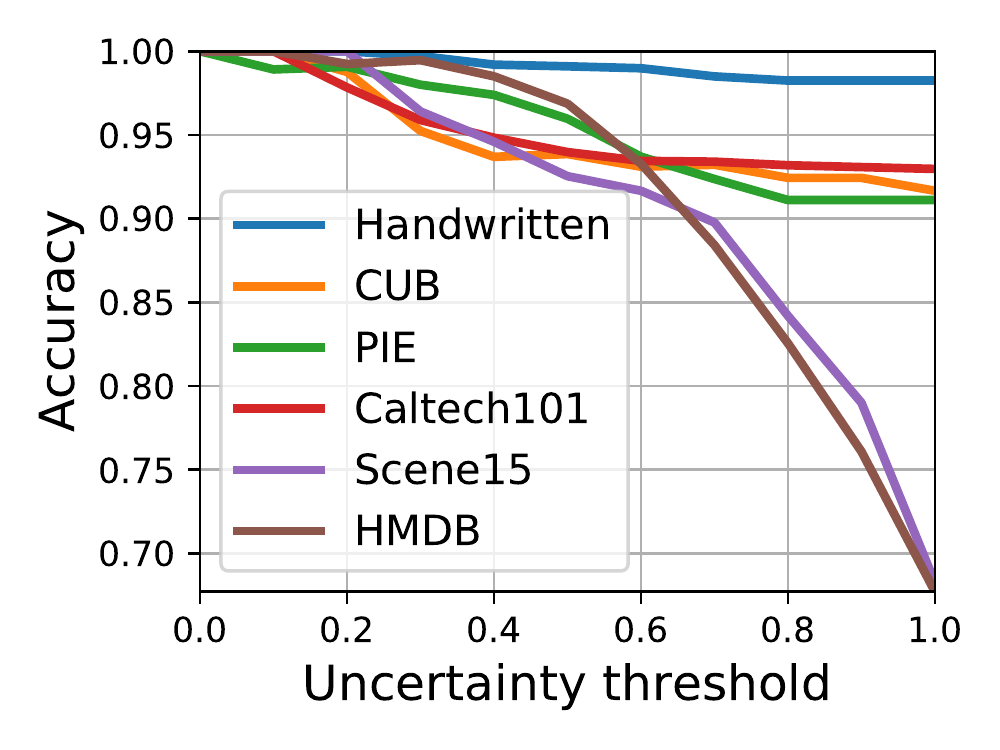}
\centering
\caption{Accuracy with uncertainty thresholding.}
\label{fig:threshold}
\end{wrapfigure}
\textbf{Compared methods}. We compare the proposed method with following models: (a) MCDO (monte carlo dropout) \citep{gal2015bayesian} casts dropout network training as approximate inference in a Bayesian neural network; (b) DE (deep ensemble) \citep{lakshminarayanan2017simple} is a simple, non-Bayesian method which involves training multiple deep models; (c) UA (uncertainty-aware attention) \citep{heo2018uncertainty}
generates attention weights following a Gaussian distribution with a learned mean and variance, which allows  heteroscedastic uncertainty to be captured and yields a more accurate calibration of prediction uncertainty; (d) EDL (evidential deep Learning) \citep{sensoy2018evidential} designs a predictive distribution for classification by placing a Dirichlet distribution on the class probabilities.

\subsection{Experimental Results}
\textbf{Comparison with uncertainty-based algorithms using the best view.} We first compare our algorithm with current uncertainty-based classification methods. The detailed experimental results are shown in Table~\ref{tab:addlabel}. Since most existing uncertainty-based classification methods use single-view data, we report the results of each method with the best-performing view in terms of both accuracy and AUROC \citep{hand2001simple} to comprehensively compare our method with others. As shown in Table~\ref{tab:addlabel}, our model outperforms other methods on all datasets. Taking the results on PIE and Scene15 as examples, our method improves the accuracy by about 7.6\% and 14.8\% compared to the second-best models (EDL/MCDO) in terms of accuracy respectively. Although our model is clearly more effective than single-view uncertainty-based models, it is natural to further ask - what happens if all algorithms utilize multiple views?
\begin{table}[!htbp]
  \centering
  \footnotesize
  \caption{Evaluation of the classification performance.}
    \begin{tabular}{ccccccc}
    \toprule
    Data  & Metric & MCDO  & DE    & UA    & EDL   & Ours \\
    \midrule
    \multirow{2}[1]{*}{Handwritten} & ACC   & 97.37$\pm$0.80 & 98.30$\pm$0.31 & 97.45$\pm$0.84 & 97.67$\pm$0.32 & \textbf{98.51$\pm$0.15} \\
          & AUROC & 99.70$\pm$0.07 & 99.79$\pm$0.05 & 99.67$\pm$0.10 & 99.83$\pm$0.02 & \textbf{99.97$\pm$0.00} \\
              \hline
    \multirow{2}[0]{*}{CUB} & ACC   & 89.78$\pm$0.52 & 90.19$\pm$0.51 & 89.75$\pm$1.43 & 89.50$\pm$1.17 & \textbf{91.00$\pm$0.42} \\
          & AUROC & \textbf{99.29$\pm$0.03} & 98.77$\pm$0.03 & 98.69$\pm$0.39 & 98.71$\pm$0.03 & 99.06$\pm$0.03 \\
              \hline
    \multirow{2}[0]{*}{PIE} & ACC   & 84.09$\pm$1.45 & 70.29$\pm$3.17 & 83.70$\pm$2.70 & 84.36$\pm$0.87 & \textbf{91.99$\pm$1.01} \\
          & AUROC & 98.90$\pm$0.31 & 95.71$\pm$0.88 & 98.06$\pm$0.56 & 98.74$\pm$0.17 & \textbf{99.69$\pm$0.05} \\
              \hline
    \multirow{2}[0]{*}{Caltech101} & ACC   & 91.73$\pm$0.58 & 91.60$\pm$0.82 & 92.37$\pm$0.72 & 90.84$\pm$0.56 & \textbf{92.93$\pm$0.20} \\
          & AUROC & 99.91$\pm$0.01 & \textbf{99.94$\pm$0.01} & 99.85$\pm$0.05 & 99.74$\pm$0.03 & 99.90$\pm$0.01 \\
          \hline
    \multirow{2}[1]{*}{Scene15} & ACC   & 52.96$\pm$1.17 & 39.12$\pm$0.80 & 41.20$\pm$1.34 & 46.41$\pm$0.55 & \textbf{67.74$\pm$0.36} \\
          & AUROC & 92.90$\pm$0.31 & 74.64$\pm$0.47 & 85.26$\pm$0.32 & 91.41$\pm$0.05 & \textbf{95.94$\pm$0.02} \\
         \hline
    \multirow{2}[0]{*}{HMDB} & ACC   & 52.92$\pm$1.28 & 57.93$\pm$1.02 & 53.32$\pm$1.39 & 59.88$\pm$1.19 & \textbf{65.26$\pm$0.76} \\
          & AUROC & 93.57$\pm$0.28 & 94.01$\pm$0.21 & 91.68$\pm$0.69 & 94.00$\pm$0.25 & \textbf{96.18$\pm$0.10} \\
    \bottomrule
    \end{tabular}%
  \label{tab:addlabel}%
\end{table}%

\begin{figure}[!htbp]
\centering
\subfigure[Handwritten]{
\centering
\begin{minipage}[t]{0.32\linewidth}
\centering
\includegraphics[width=1\linewidth,height=0.7\linewidth]{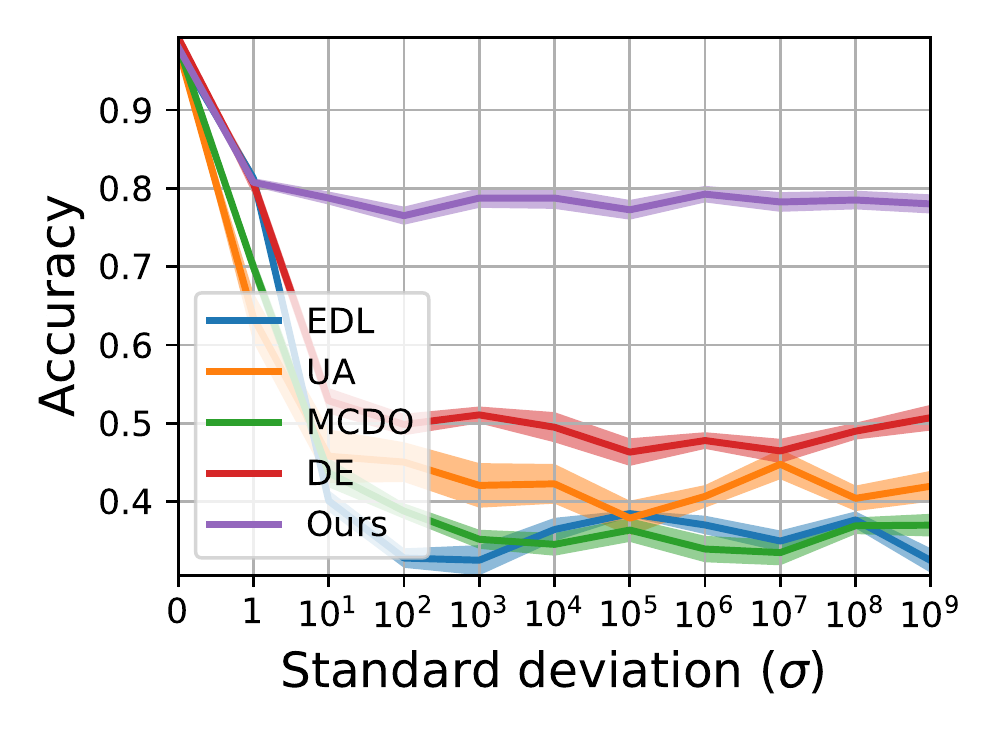}
\centering
\end{minipage}}
\subfigure[CUB]{
\begin{minipage}[t]{0.32\linewidth}
\centering
\includegraphics[width=1\linewidth,height=0.7\linewidth]{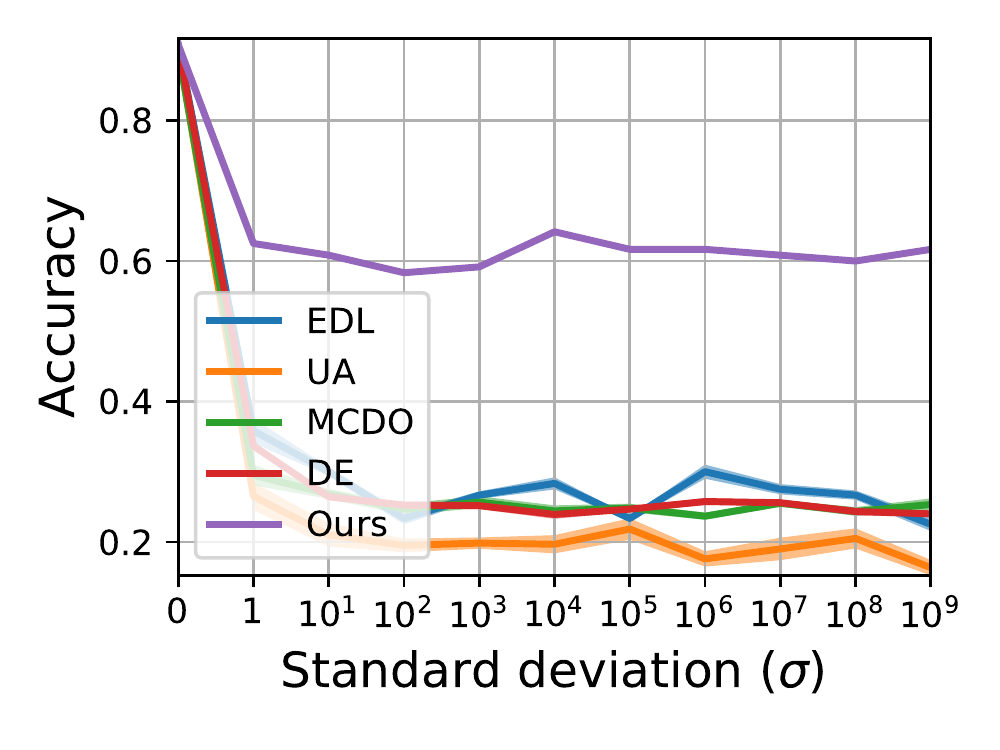}
\end{minipage}}
\centering
\subfigure[PIE]{
\begin{minipage}[t]{0.32\linewidth}
\centering
\includegraphics[width=1\linewidth,height=0.7\linewidth]{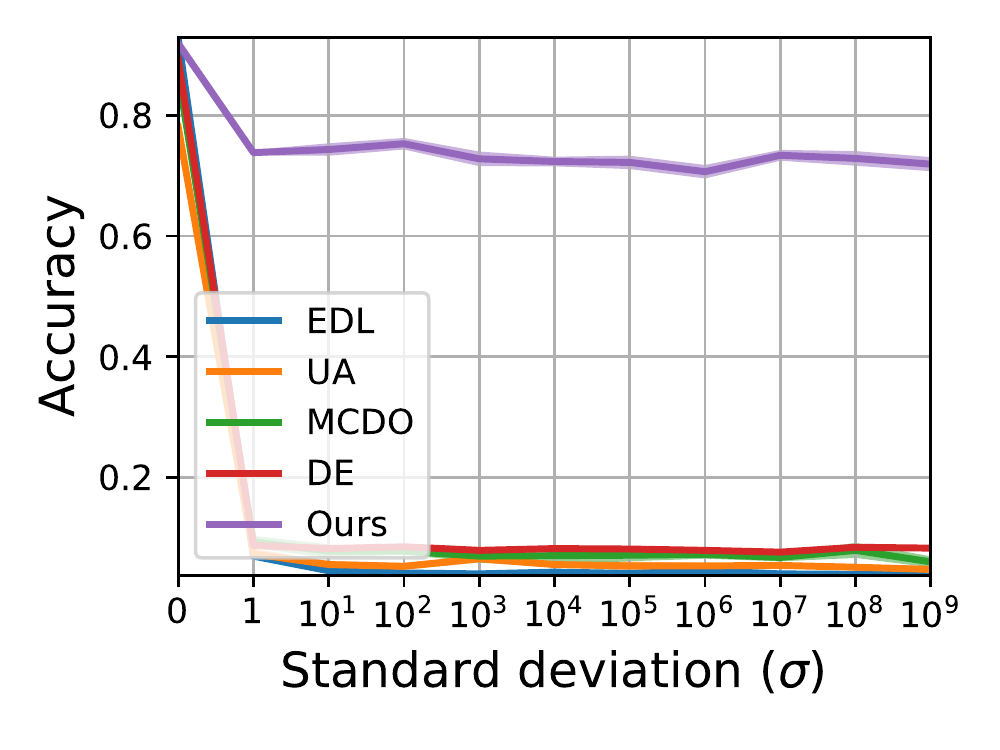}
\end{minipage}}
\subfigure[Caltech101]{
\begin{minipage}[t]{0.32\linewidth}
\centering
\includegraphics[width=1\linewidth,height=0.7\linewidth]{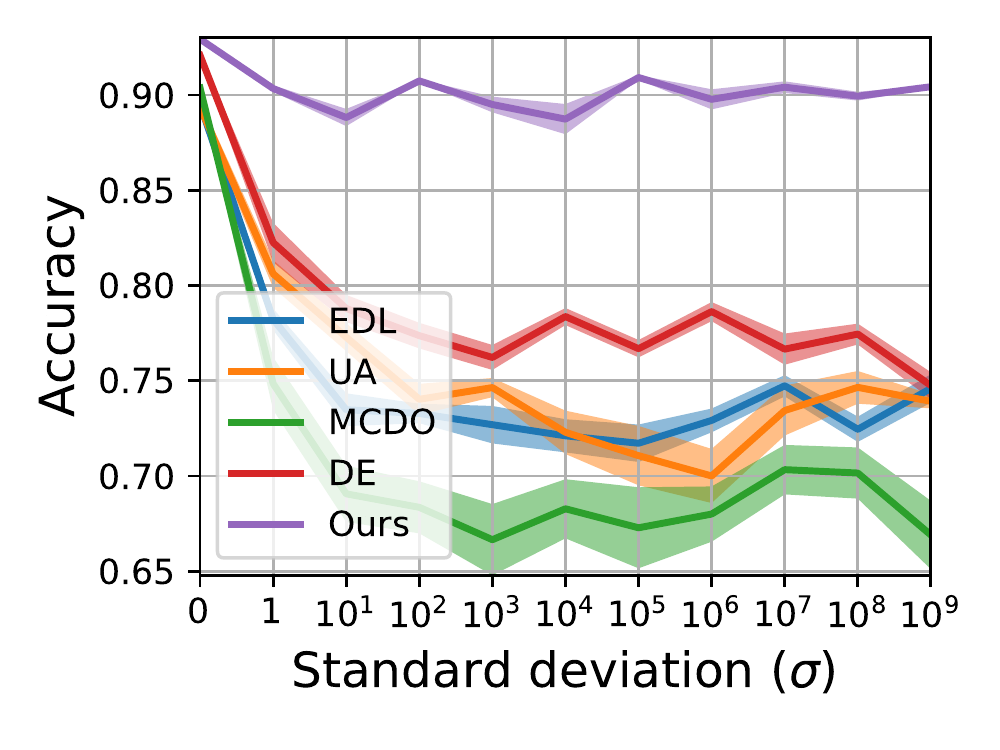}
\end{minipage}}
\centering
\centering
\subfigure[Scene15]{
\begin{minipage}[t]{0.32\linewidth}
\centering
\includegraphics[width=1\linewidth,height=0.7\linewidth]{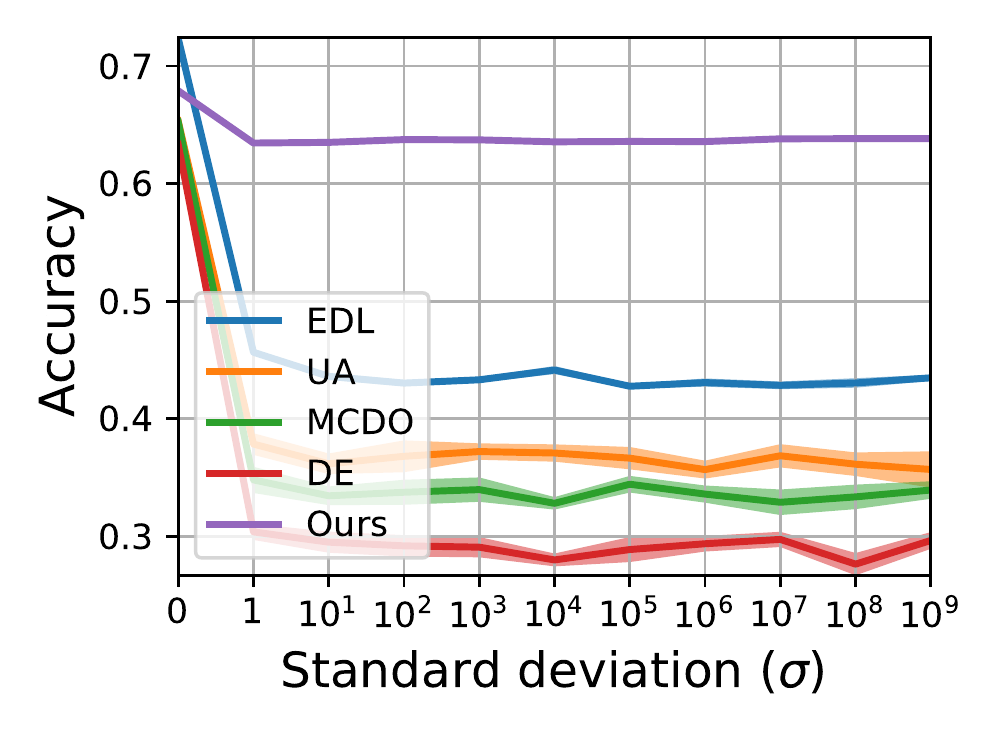}
\end{minipage}}
\subfigure[HMDB]{
\begin{minipage}[t]{0.32\linewidth}
\centering
\includegraphics[width=1\linewidth,height=0.7\linewidth]{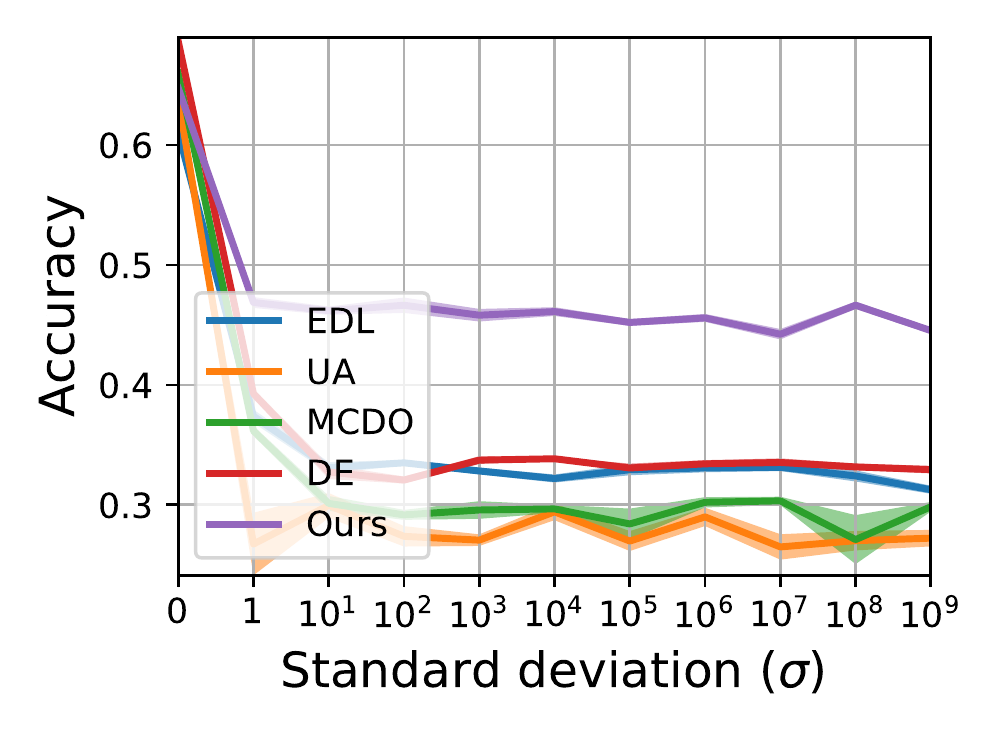}
\end{minipage}}
\centering
\caption{Performance comparison on multi-view data with different levels of noise.}
\label{fig:linechart}
\end{figure}

\textbf{Comparison with uncertainty-based algorithms using multiple views.} To further validate the effectiveness of our model in integrating different various views, we concatenate the original features of multiple views for all comparison methods.
We add Gaussian noise with different levels of standard deviations ($\sigma$) to half of the views. The comparison results are shown in Fig.~\ref{fig:linechart}. As can be observed that when the data is free of noise, our method can achieve competitive results. After introducing noise to the data, the accuracy of all the comparison methods significantly decreases. Fortunately, benefiting from the uncertainty-based fusion, the proposed method is aware of the view-specific noise and thus achieves impressive results on all datasets. Therefore, the effectiveness for both clean and noisy multi-view data is well validated. However, it will be more convincing to explicitly investigate the performance in uncertainty estimation.

\begin{figure}[!t]
\centering
\subfigure[Handwritten]{
\centering
\begin{minipage}[t]{0.3\linewidth}
\centering
\includegraphics[width=0.9\linewidth,height=0.744\linewidth]{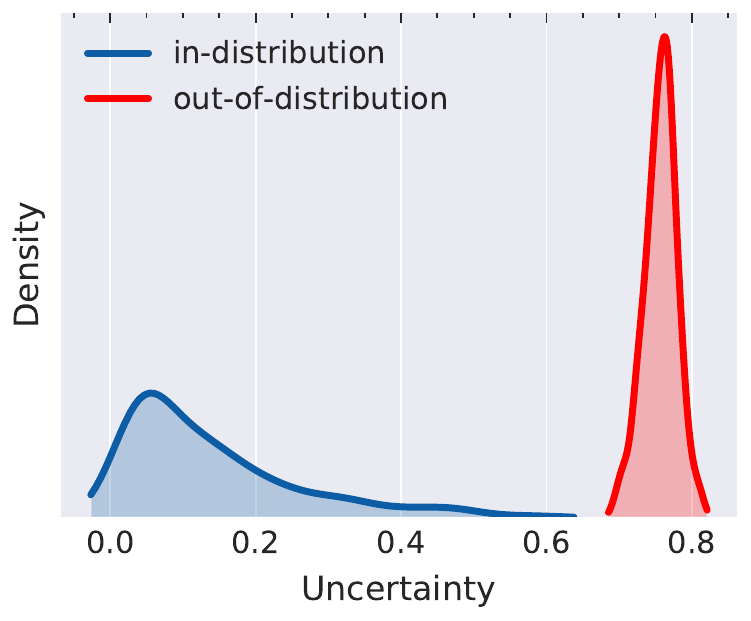}
\centering
\end{minipage}}
\subfigure[CUB]{
\begin{minipage}[t]{0.3\linewidth}
\centering
\includegraphics[width=0.9\linewidth,height=0.744\linewidth]{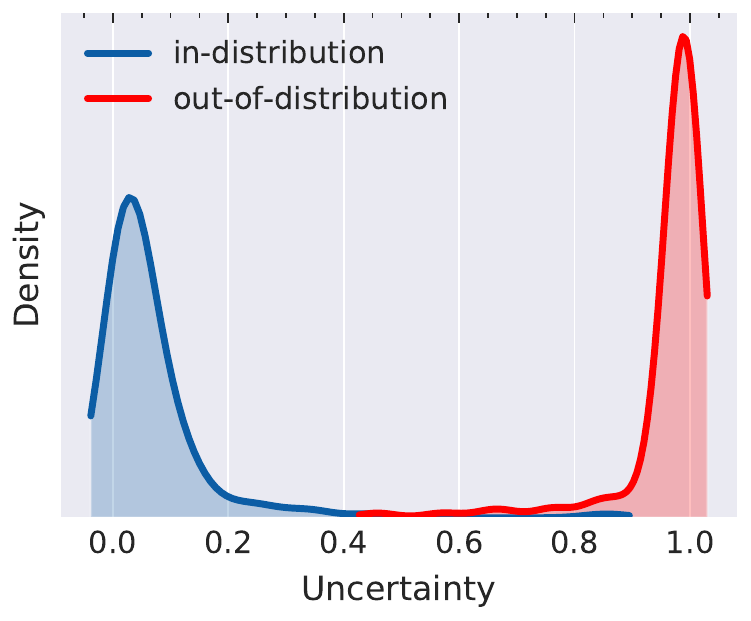}
\end{minipage}}
\centering
\subfigure[PIE]{
\begin{minipage}[t]{0.3\linewidth}
\centering
\includegraphics[width=0.9\linewidth,height=0.744\linewidth]{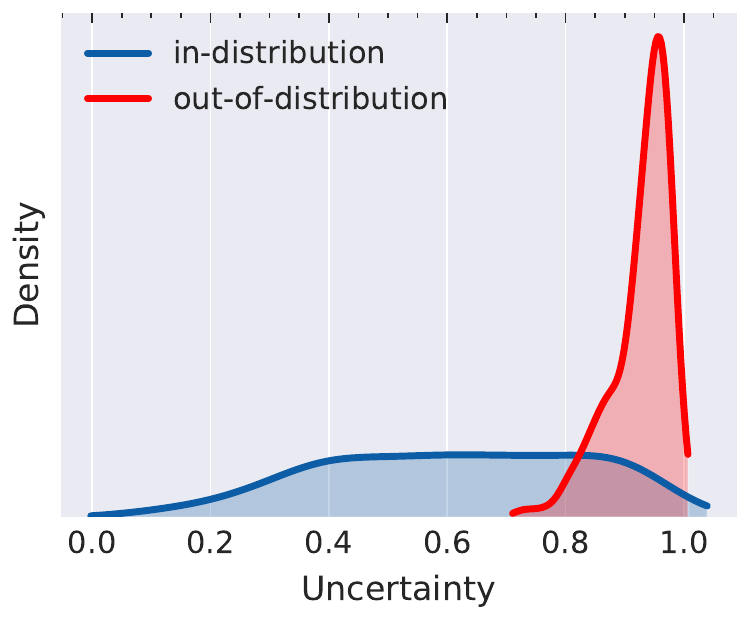}
\end{minipage}}
\centering
\subfigure[Caltech101]{
\begin{minipage}[t]{0.3\linewidth}
\centering
\includegraphics[width=0.9\linewidth,height=0.744\linewidth]{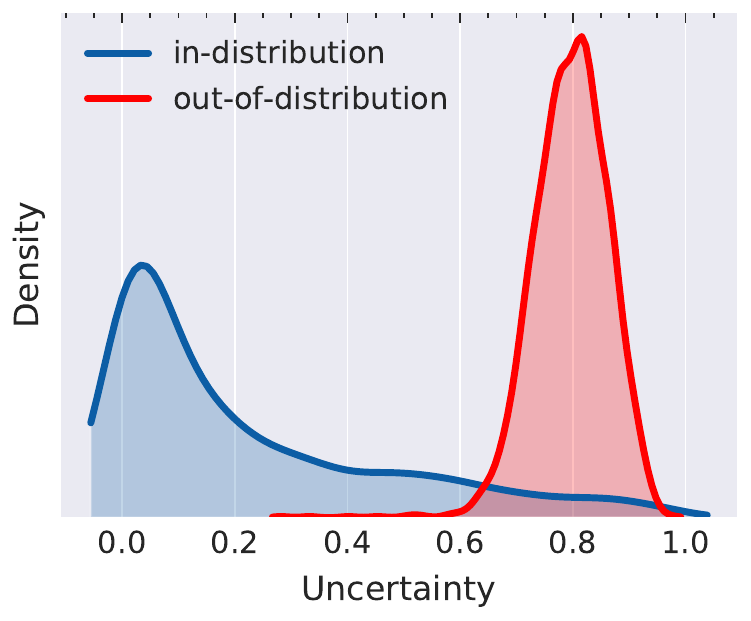}
\end{minipage}}
\centering
\subfigure[Scene15]{
\begin{minipage}[t]{0.3\linewidth}
\centering
\includegraphics[width=0.9\linewidth,height=0.744\linewidth]{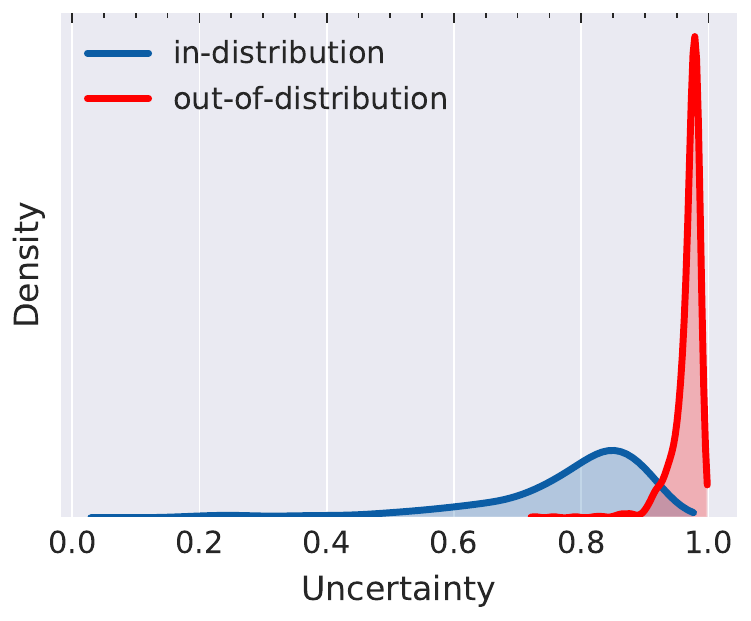}
\end{minipage}}
\centering
\subfigure[HMDB]{
\begin{minipage}[t]{0.3\linewidth}
\centering
\includegraphics[width=0.9\linewidth,height=0.744\linewidth]{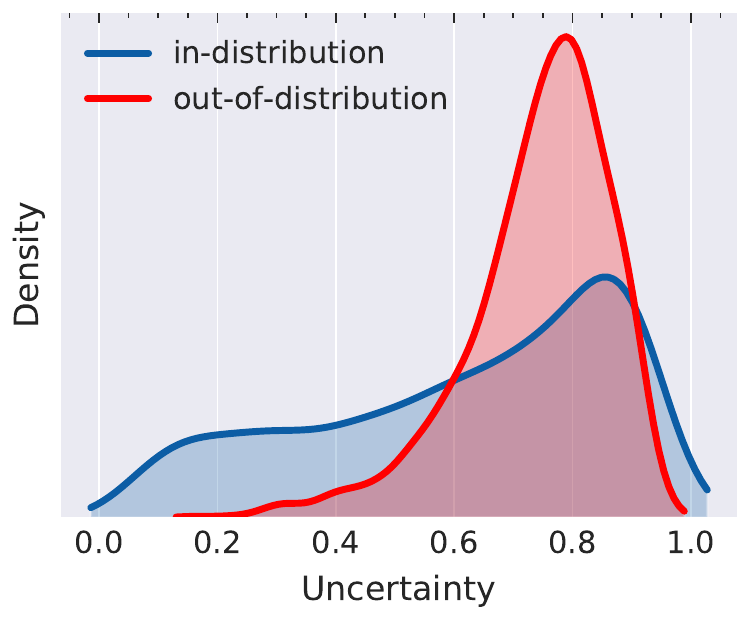}
\end{minipage}}
\caption{Density of uncertainty.}
\label{fig:noisedensity}
\end{figure}

\textbf{Uncertainty estimation.} To evaluate the uncertainty estimation, we visualize the distribution of in-/out-of-distribution samples in terms of uncertainty. We consider the original samples as in-distribution data, while the samples with Gaussian noise are viewed as out-of-distribution data. Specifically, we add Gaussian noise with the fixed level of standard deviations ($\sigma = 10$) to $50\%$ of the test samples. The experimental results are shown in Fig.~\ref{fig:noisedensity}. According to the results, the following observations are drawn: (1) Datasets with higher classification accuracy (\emph{e.g.}, Handwritten) are usually associated with lower uncertainty for the in-distribution samples. (2) In contrast, datasets with lower accuracy are usually associated with higher uncertainty for the in-distribution samples. (3) Much higher uncertainties are usually estimated for out-of-distribution samples on all datasets. These observations imply the reasonability of our model in estimating uncertainty, since it can facilitate discrimination between these classes. Fig.~\ref{fig:threshold} shows that our algorithm provides much more accurate predictions as the prediction uncertainty decreases. This implies that trusted decisions are supported based on the output (classification and its corresponding uncertainty) of our model.

\section{Conclusion}
In this work, we propose a novel trusted multi-view classification (TMC) algorithm which, based on the Dempster-Shafer evidence theory, can produce trusted classification decisions on multi-view data. Our algorithm focuses on decision-making by fusing the uncertainty of multiple views, which is essential for making trusted decisions. The TMC model can accurately identify the views which are risky for decision making, and exploits informative views in the final decision. Furthermore, our model can produce the uncertainty of a current decision while making the final classification, providing intepretability. The empirical results validate the effectiveness of the proposed algorithm in classification accuracy and out-of-distribution identification.

\subsubsection*{Acknowledgments}
This work was supported in part by National Natural Science Foundation of China (No. 61976151, No. 61732011), and the Natural Science Foundation of Tianjin of China (No. 19JCYBJC15200).
\bibliography{iclr2021_conference}

\begin{thebibliography}{53}
\providecommand{\natexlab}[1]{#1}
\providecommand{\url}[1]{\texttt{#1}}
\expandafter\ifx\csname urlstyle\endcsname\relax
  \providecommand{\doi}[1]{doi: #1}\else
  \providecommand{\doi}{doi: \begingroup \urlstyle{rm}\Url}\fi

\bibitem[Akaho(2006)]{akaho2006kernel}
Shotaro Akaho.
\newblock A kernel method for canonical correlation analysis.
\newblock \emph{arXiv preprint cs/0609071}, 2006.

\bibitem[Andrew et~al.(2013)Andrew, Arora, Bilmes, and Livescu]{andrew2013deep}
Galen Andrew, Raman Arora, Jeff Bilmes, and Karen Livescu.
\newblock Deep canonical correlation analysis.
\newblock In \emph{International Conference on Machine Learning}, pp.\
  1247--1255, 2013.

\bibitem[Bachman et~al.(2019)Bachman, Hjelm, and
  Buchwalter]{bachman2019learning}
Philip Bachman, R~Devon Hjelm, and William Buchwalter.
\newblock Learning representations by maximizing mutual information across
  views.
\newblock In \emph{Advances in Neural Information Processing Systems}, pp.\
  15535--15545, 2019.

\bibitem[Bernardo \& Smith(2009)Bernardo and Smith]{bernardo2009bayesian}
Jos{\'e}~M Bernardo and Adrian~FM Smith.
\newblock \emph{Bayesian theory}, volume 405.
\newblock John Wiley \& Sons, 2009.

\bibitem[Bian et~al.(2017)Bian, Gan, Liu, Li, Long, Li, Qi, Zhou, Wen, and
  Lin]{bian2017revisiting}
Yunlong Bian, Chuang Gan, Xiao Liu, Fu~Li, Xiang Long, Yandong Li, Heng Qi, Jie
  Zhou, Shilei Wen, and Yuanqing Lin.
\newblock Revisiting the effectiveness of off-the-shelf temporal modeling
  approaches for large-scale video classification.
\newblock \emph{arXiv preprint arXiv:1708.03805}, 2017.

\bibitem[Bishop(2006)]{bishop2006pattern}
Christopher~M Bishop.
\newblock \emph{Pattern recognition and machine learning}.
\newblock springer, 2006.

\bibitem[Blundell et~al.(2015)Blundell, Cornebise, Kavukcuoglu, and
  Wierstra]{blundell2015weight}
Charles Blundell, Julien Cornebise, Koray Kavukcuoglu, and Daan Wierstra.
\newblock Weight uncertainty in neural network.
\newblock In \emph{International Conference on Machine Learning}, pp.\
  1613--1622, 2015.

\bibitem[Chen et~al.(2020)Chen, Kornblith, Norouzi, and Hinton]{chen2020simple}
Ting Chen, Simon Kornblith, Mohammad Norouzi, and Geoffrey Hinton.
\newblock A simple framework for contrastive learning of visual
  representations.
\newblock \emph{arXiv preprint arXiv:2002.05709}, 2020.

\bibitem[Dempster(1967)]{dempster1967upper}
AP~Dempster.
\newblock Upper and lower probabilities induced by a multivalued mapping.
\newblock \emph{The Annals of Mathematical Statistics}, pp.\  325--339, 1967.

\bibitem[Dempster(1968)]{dempster1968generalization}
Arthur~P Dempster.
\newblock A generalization of bayesian inference.
\newblock \emph{Journal of the Royal Statistical Society: Series B
  (Methodological)}, 30\penalty0 (2):\penalty0 205--232, 1968.

\bibitem[Denker \& LeCun(1991)Denker and LeCun]{denker1991transforming}
John~S Denker and Yann LeCun.
\newblock Transforming neural-net output levels to probability distributions.
\newblock In \emph{Advances in Neural Information Processing Systems}, pp.\
  853--859, 1991.

\bibitem[Devlin et~al.(2018)Devlin, Chang, Lee, and Toutanova]{devlin2018bert}
Jacob Devlin, Ming-Wei Chang, Kenton Lee, and Kristina Toutanova.
\newblock Bert: Pre-training of deep bidirectional transformers for language
  understanding.
\newblock \emph{arXiv preprint arXiv:1810.04805}, 2018.

\bibitem[Fei-Fei \& Perona(2005)Fei-Fei and Perona]{fei2005bayesian}
Li~Fei-Fei and Pietro Perona.
\newblock A bayesian hierarchical model for learning natural scene categories.
\newblock In \emph{2005 IEEE Computer Society Conference on Computer Vision and
  Pattern Recognition (CVPR'05)}, volume~2, pp.\  524--531. IEEE, 2005.

\bibitem[Fei-Fei et~al.(2004)Fei-Fei, Fergus, and Perona]{fei2004learning}
Li~Fei-Fei, Rob Fergus, and Pietro Perona.
\newblock Learning generative visual models from few training examples: An
  incremental bayesian approach tested on 101 object categories.
\newblock In \emph{2004 Conference on Computer Vision and Pattern Recognition
  workshop}, pp.\  178--178. IEEE, 2004.

\bibitem[Frigyik et~al.(2010)Frigyik, Kapila, and
  Gupta]{frigyik2010introduction}
Bela~A Frigyik, Amol Kapila, and Maya~R Gupta.
\newblock Introduction to the dirichlet distribution and related processes.
\newblock \emph{Department of Electrical Engineering, University of Washignton,
  UWEETR-2010-0006}, \penalty0 (0006):\penalty0 1--27, 2010.

\bibitem[Gal \& Ghahramani(2015)Gal and Ghahramani]{gal2015bayesian}
Yarin Gal and Zoubin Ghahramani.
\newblock Bayesian convolutional neural networks with bernoulli approximate
  variational inference.
\newblock \emph{arXiv preprint arXiv:1506.02158}, 2015.

\bibitem[Gal \& Ghahramani(2016)Gal and Ghahramani]{gal2016dropout}
Yarin Gal and Zoubin Ghahramani.
\newblock Dropout as a bayesian approximation: Representing model uncertainty
  in deep learning.
\newblock In \emph{International Conference on Machine Learning}, pp.\
  1050--1059, 2016.

\bibitem[Graves(2011)]{graves2011practical}
Alex Graves.
\newblock Practical variational inference for neural networks.
\newblock In \emph{Advances in Neural Information Processing Systems}, pp.\
  2348--2356, 2011.

\bibitem[Hand \& Till(2001)Hand and Till]{hand2001simple}
David~J Hand and Robert~J Till.
\newblock A simple generalisation of the area under the roc curve for multiple
  class classification problems.
\newblock \emph{Machine learning}, 45\penalty0 (2):\penalty0 171--186, 2001.

\bibitem[Hassani \& Khasahmadi(2020)Hassani and
  Khasahmadi]{hassani2020contrastive}
Kaveh Hassani and Amir~Hosein Khasahmadi.
\newblock Contrastive multi-view representation learning on graphs.
\newblock \emph{arXiv preprint arXiv:2006.05582}, 2020.

\bibitem[He et~al.(2016)He, Zhang, Ren, and Sun]{he2016deep}
Kaiming He, Xiangyu Zhang, Shaoqing Ren, and Jian Sun.
\newblock Deep residual learning for image recognition.
\newblock In \emph{Proceedings of the IEEE Conference on Computer Vision and
  Pattern Recognition}, pp.\  770--778, 2016.

\bibitem[Heo et~al.(2018)Heo, Lee, Kim, Lee, Kim, Yang, and
  Hwang]{heo2018uncertainty}
Jay Heo, Hae~Beom Lee, Saehoon Kim, Juho Lee, Kwang~Joon Kim, Eunho Yang, and
  Sung~Ju Hwang.
\newblock Uncertainty-aware attention for reliable interpretation and
  prediction.
\newblock In \emph{Advances in Neural Information Processing Systems}, pp.\
  909--918, 2018.

\bibitem[Hotelling(1992)]{hotelling1992relations}
Harold Hotelling.
\newblock Relations between two sets of variates.
\newblock In \emph{Breakthroughs in statistics}, pp.\  162--190. Springer,
  1992.

\bibitem[J{\o}sang(2018)]{jsang2018subjective}
Audun J{\o}sang.
\newblock \emph{Subjective Logic: A formalism for reasoning under uncertainty}.
\newblock Springer Publishing Company, Incorporated, 2018.

\bibitem[J{\o}sang \& Hankin(2012)J{\o}sang and
  Hankin]{josang2012interpretation}
Audun J{\o}sang and Robin Hankin.
\newblock Interpretation and fusion of hyper opinions in subjective logic.
\newblock In \emph{2012 15th International Conference on Information Fusion},
  pp.\  1225--1232. IEEE, 2012.

\bibitem[Kendall et~al.(2018)Kendall, Gal, and Cipolla]{kendall2018multi}
Alex Kendall, Yarin Gal, and Roberto Cipolla.
\newblock Multi-task learning using uncertainty to weigh losses for scene
  geometry and semantics.
\newblock In \emph{Proceedings of the IEEE Conference on Computer Vision and
  Pattern Recognition}, pp.\  7482--7491, 2018.

\bibitem[Kiela et~al.(2018)Kiela, Grave, Joulin, and
  Mikolov]{kiela2018efficient}
Douwe Kiela, Edouard Grave, Armand Joulin, and Tomas Mikolov.
\newblock Efficient large-scale multi-modal classification.
\newblock \emph{arXiv preprint arXiv:1802.02892}, 2018.

\bibitem[Kiela et~al.(2019)Kiela, Bhooshan, Firooz, and
  Testuggine]{kiela2019supervised}
Douwe Kiela, Suvrat Bhooshan, Hamed Firooz, and Davide Testuggine.
\newblock Supervised multimodal bitransformers for classifying images and text.
\newblock \emph{arXiv preprint arXiv:1909.02950}, 2019.

\bibitem[Kingma \& Ba(2014)Kingma and Ba]{kingma2014adam}
Diederik~P Kingma and Jimmy Ba.
\newblock Adam: A method for stochastic optimization.
\newblock \emph{arXiv preprint arXiv:1412.6980}, 2014.

\bibitem[Kuehne et~al.(2011)Kuehne, Jhuang, Garrote, Poggio, and
  Serre]{kuehne2011hmdb}
Hildegard Kuehne, Hueihan Jhuang, Est{\'\i}baliz Garrote, Tomaso Poggio, and
  Thomas Serre.
\newblock Hmdb: a large video database for human motion recognition.
\newblock In \emph{2011 International Conference on Computer Vision}, pp.\
  2556--2563. IEEE, 2011.

\bibitem[Lakshminarayanan et~al.(2017)Lakshminarayanan, Pritzel, and
  Blundell]{lakshminarayanan2017simple}
Balaji Lakshminarayanan, Alexander Pritzel, and Charles Blundell.
\newblock Simple and scalable predictive uncertainty estimation using deep
  ensembles.
\newblock In \emph{Advances in Neural Information Processing Systems}, pp.\
  6402--6413, 2017.

\bibitem[MacKay(1992{\natexlab{a}})]{mackay1992bayesian}
David~JC MacKay.
\newblock \emph{Bayesian methods for adaptive models}.
\newblock PhD thesis, California Institute of Technology, 1992{\natexlab{a}}.

\bibitem[MacKay(1992{\natexlab{b}})]{mackay1992practical}
David~JC MacKay.
\newblock A practical bayesian framework for backpropagation networks.
\newblock \emph{Neural computation}, 4\penalty0 (3):\penalty0 448--472,
  1992{\natexlab{b}}.

\bibitem[Moon et~al.(2020)Moon, Kim, Shin, and Hwang]{moon2020confidence}
Jooyoung Moon, Jihyo Kim, Younghak Shin, and Sangheum Hwang.
\newblock Confidence-aware learning for deep neural networks.
\newblock In \emph{International Conference on Machine Learning}, 2020.

\bibitem[Neal(2012)]{neal2012bayesian}
Radford~M Neal.
\newblock \emph{Bayesian learning for neural networks}, volume 118.
\newblock Springer Science \& Business Media, 2012.

\bibitem[Perrin et~al.(2009)Perrin, Fagan, and Holtzman]{perrin2009multimodal}
Richard~J Perrin, Anne~M Fagan, and David~M Holtzman.
\newblock Multimodal techniques for diagnosis and prognosis of alzheimer's
  disease.
\newblock \emph{Nature}, 461\penalty0 (7266):\penalty0 916--922, 2009.

\bibitem[Ranganath et~al.(2014)Ranganath, Gerrish, and
  Blei]{ranganath2014black}
Rajesh Ranganath, Sean Gerrish, and David Blei.
\newblock Black box variational inference.
\newblock In \emph{Artificial Intelligence and Statistics}, pp.\  814--822.
  PMLR, 2014.

\bibitem[Sensoy et~al.(2018)Sensoy, Kaplan, and Kandemir]{sensoy2018evidential}
Murat Sensoy, Lance Kaplan, and Melih Kandemir.
\newblock Evidential deep learning to quantify classification uncertainty.
\newblock In \emph{Advances in Neural Information Processing Systems}, pp.\
  3179--3189, 2018.

\bibitem[Sentz et~al.(2002)Sentz, Ferson, et~al.]{sentz2002combination}
Kari Sentz, Scott Ferson, et~al.
\newblock \emph{Combination of evidence in Dempster-Shafer theory}, volume
  4015.
\newblock Citeseer, 2002.

\bibitem[Shafer(1976)]{shafer1976mathematical}
Glenn Shafer.
\newblock \emph{A mathematical theory of evidence}, volume~42.
\newblock Princeton university press, 1976.

\bibitem[Silberman et~al.(2012)Silberman, Hoiem, Kohli, and
  Fergus]{silberman2012indoor}
Nathan Silberman, Derek Hoiem, Pushmeet Kohli, and Rob Fergus.
\newblock Indoor segmentation and support inference from rgbd images.
\newblock In \emph{European Conference on Computer vision}, pp.\  746--760.
  Springer, 2012.

\bibitem[Srivastava et~al.(2014)Srivastava, Hinton, Krizhevsky, Sutskever, and
  Salakhutdinov]{srivastava2014dropout}
Nitish Srivastava, Geoffrey Hinton, Alex Krizhevsky, Ilya Sutskever, and Ruslan
  Salakhutdinov.
\newblock Dropout: a simple way to prevent neural networks from overfitting.
\newblock \emph{The Journal of Machine Learning Research}, 15\penalty0
  (1):\penalty0 1929--1958, 2014.

\bibitem[Sui et~al.(2018)Sui, Qi, van Erp, Bustillo, Jiang, Lin, Turner,
  Damaraju, Mayer, Cui, et~al.]{sui2018multimodal}
Jing Sui, Shile Qi, Theo~GM van Erp, Juan Bustillo, Rongtao Jiang, Dongdong
  Lin, Jessica~A Turner, Eswar Damaraju, Andrew~R Mayer, Yue Cui, et~al.
\newblock Multimodal neuromarkers in schizophrenia via cognition-guided mri
  fusion.
\newblock \emph{Nature communications}, 9\penalty0 (1):\penalty0 1--14, 2018.

\bibitem[Tian et~al.(2019)Tian, Krishnan, and Isola]{tian2019contrastive}
Yonglong Tian, Dilip Krishnan, and Phillip Isola.
\newblock Contrastive multiview coding.
\newblock \emph{arXiv preprint arXiv:1906.05849}, 2019.

\bibitem[van Amersfoort et~al.(2020)van Amersfoort, Smith, Teh, and
  Gal]{van2020uncertainty}
Joost van Amersfoort, Lewis Smith, Yee~Whye Teh, and Yarin Gal.
\newblock Uncertainty estimation using a single deep deterministic neural
  network.
\newblock In \emph{International Conference on Machine Learning}, 2020.

\bibitem[Wah et~al.(2011)Wah, Branson, Welinder, Perona, and
  Belongie]{wah2011caltech}
Catherine Wah, Steve Branson, Peter Welinder, Pietro Perona, and Serge
  Belongie.
\newblock The caltech-ucsd birds-200-2011 dataset.
\newblock 2011.

\bibitem[Wang(2007)]{wang2007variational}
Chong Wang.
\newblock Variational bayesian approach to canonical correlation analysis.
\newblock \emph{IEEE Transactions on Neural Networks}, 18\penalty0
  (3):\penalty0 905--910, 2007.

\bibitem[Wang et~al.(2015{\natexlab{a}})Wang, Arora, Livescu, and
  Bilmes]{wang2015deep}
Weiran Wang, Raman Arora, Karen Livescu, and Jeff Bilmes.
\newblock On deep multi-view representation learning.
\newblock In \emph{International Conference on Machine Learning}, pp.\
  1083--1092, 2015{\natexlab{a}}.

\bibitem[Wang et~al.(2016)Wang, Yan, Lee, and Livescu]{wang2016deep}
Weiran Wang, Xinchen Yan, Honglak Lee, and Karen Livescu.
\newblock Deep variational canonical correlation analysis.
\newblock \emph{arXiv preprint arXiv:1610.03454}, 2016.

\bibitem[Wang et~al.(2020)Wang, Tran, and Feiszli]{wang2020makes}
Weiyao Wang, Du~Tran, and Matt Feiszli.
\newblock What makes training multi-modal classification networks hard?
\newblock In \emph{Proceedings of the IEEE/CVF Conference on Computer Vision
  and Pattern Recognition}, pp.\  12695--12705, 2020.

\bibitem[Wang et~al.(2015{\natexlab{b}})Wang, Kumar, Thome, Cord, and
  Precioso]{wang2015recipe}
Xin Wang, Devinder Kumar, Nicolas Thome, Matthieu Cord, and Frederic Precioso.
\newblock Recipe recognition with large multimodal food dataset.
\newblock In \emph{2015 IEEE International Conference on Multimedia \& Expo
  Workshops (ICMEW)}, pp.\  1--6. IEEE, 2015{\natexlab{b}}.

\bibitem[Zhang et~al.(2019)Zhang, Han, Fu, Zhou, Hu, et~al.]{zhang2019cpm}
Changqing Zhang, Zongbo Han, Huazhu Fu, Joey~Tianyi Zhou, Qinghua Hu, et~al.
\newblock Cpm-nets: Cross partial multi-view networks.
\newblock In \emph{Advances in Neural Information Processing Systems}, pp.\
  559--569, 2019.

\bibitem[Zhang et~al.(2017)Zhang, Patel, and Chellappa]{zhang2017hierarchical}
Heng Zhang, Vishal~M Patel, and Rama Chellappa.
\newblock Hierarchical multimodal metric learning for multimodal
  classification.
\newblock In \emph{Proceedings of the IEEE Conference on Computer Vision and
  Pattern Recognition}, pp.\  3057--3065, 2017.

\end{thebibliography}
\bibliographystyle{iclr2021_conference}
\clearpage
\appendix
\section{Appendix}
\subsection{Definition of Dirichlet distribution}
\begin{definition} (\textbf{Dirichlet distribution})
The Dirichlet distribution is parameterized by its concentration $K$ parameters $\mathbf{\boldsymbol{\alpha}}=[\alpha_1, \ldots, \alpha_K]$. \textcolor{black}{The probability density function of the Dirichlet distribution is given by}
\begin{equation}D(\mathbf{p} \mid \boldsymbol{\alpha})=\left\{\begin{array}{ll}
\frac{1}{B(\boldsymbol{\alpha})} \prod_{i=1}^{K} p_{i}^{\alpha_{i}-1} & \text { for } \mathbf{p} \in \mathcal{S}_{K} ,\\
0 & \text { otherwise ,}
\end{array}\right.\end{equation}
where $\mathcal{S}_{K}$ is the $K$-dimensional unit simplex, defined as
\begin{equation}
\mathcal{S}_{K}=\left\{\mathbf{p} \mid \sum_{i=1}^{K} p_{i}=1 \text { and } 0 \leq p_{1}, \ldots, p_{K} \leq 1\right\},
\end{equation}
and $B(\boldsymbol{\alpha})$ is the K-dimensional multinomial beta function.
\label{def:dirichlet}
\end{definition}
\subsection{Algorithm for trust Multi-View Classification}
\begin{algorithm}[h]
\SetAlgoLined
\caption{Algorithm for Trusted Multi-View Classification}
\textbf{/*Training*/}\\
\KwIn{Multi-view dataset: $\mathcal{D} = \{\{\mathbf{X}_n^v\}_{v=1}^{V}, y_n\}_{n=1}^N$.}
\textbf{Initialize:} {Initialize the parameters of the neural network}.\\
\While{not converged}{
\For{$v=1:V$ }
{
$\mathbf{e}^v\leftarrow$ evidential network output;\\
$\mathbf{\boldsymbol{\alpha}}^v\leftarrow \mathbf{e}^v + 1$ ;\\
Obtain opinion $\mathcal{M}^v$ with Eq.~\ref{eq:sl};\\
}
Obtain joint opinion $\mathcal{M}$ with Eq.~\ref{eq:fusion2};\\
Obtain $\boldsymbol{\alpha}$ with Eq.~\ref{eq:sl2};\\
%Obtain the overall loss with Eq.~\ref{eq:loss}, $\boldsymbol{\alpha}$ and $\{\mathbf{\boldsymbol{\alpha}}^v\}_{v=1}^{V}$;\\
Obtain the overall loss by updating $\boldsymbol{\alpha}$ and $\{\mathbf{\boldsymbol{\alpha}}^v\}_{v=1}^{V}$ in Eq.~\ref{eq:loss};\\
Update the networks with gradient descent according to Eq.~\ref{eq:loss};
}
\KwOut{networks parameters.}%,\\
\textbf{/*Test*/}\\
Calculate the joint belief mass and the uncertainty mass.
\label{alg:alg1}
\end{algorithm}
\subsection{Details of the datasets}
{To better evaluate our model, we conduct experiments on the following six real-world datasets. Details of these datasets are as follows: }
\begin{itemize}
    \item {\textbf{1) Handwritten} consists of 2000 samples of 10 classes from digit `0' to `9' with 200 samples per class, where six different types of descriptors are used as multiple views.}
    \item {\textbf{2) CUB} consists of 11788 bird images associated with text descriptions from 200 different categories of birds. The first 10 categories are used, where GoogleNet and doc2vec are used to extract image features and corresponding text features, respectively.}
    \item {\textbf{3) PIE} consists of 680 facial images of 68 subjects. Three types of features including intensity, LBP and Gabor are extracted.}
    \item {\textbf{4) HMDB} is one of the largest human action recognition dataset. There are 6718 samples of 51 categories of actions, where HOG and MBH features are extracted.}
    \item {\textbf{5) Caltech101} consists of 8677 images from 101 classes. We extract two types of deep features with DECAF and VGG19 respectively.}
    \item {\textbf{6) Scene15} consists of 4485 images from 15 indoor and outdoor scene categories. Three types features (GIST, PHOG and LBP) are extracted as multiple views.}
\end{itemize}
{\subsection{Comparison with CCA-based algorithms}}
{We compared our method with the CCA-based algorithms. Specifically, we employ CCA-based methods to obtain the latent representations and then the linear SVM classifier is used for classification. The following CCA-based algorithms are used as baselines. (1) CCA \citep{hotelling1992relations} is a classical algorithm for seeking the correlations between multiple types of features. (2) DCCA \citep{andrew2013deep} obtains the correlations through deep neural networks. (3) DCCAE \citep{wang2015deep} employs autoencoders for seeking the common representations. (4) BCCA \citep{wang2007variational} presents a Bayesian model selection algorithm for CCA based on a probabilistic interpretation. Due to  randomness (e.g., training/testing partition and optimization) involved, for each method, we run 30 times and report its mean and standard deviation in terms of classification accuracy. The experimental results are shown in Table \ref{tab:cca}. On all datasets, our method consistently achieves better performance compared with these CCA-based algorithms. Note that our method is quite different from the CCA-based methods. Ours is a classification model while most CCA-based methods are unsupervised representation learning models. Meanwhile, to the best of our knowledge, existing CCA-based algorithms are unable to provide trusted decisions.}

\begin{table}[htbp]
\footnotesize
 \centering
 {
 \caption{Comparison with CCA-based algorithms.}
   \begin{tabular}{ccccccc}
   \toprule
   Data & CCA   & DCCA  & DCCAE & BCCA  & Ours \\
   \midrule
   {Handwritten} & 97.25$\pm$0.01 & 97.55$\pm$0.38 & 97.35$\pm$0.35 & 95.75$\pm$1.23 & 98.51$\pm$0.13 \\
   \midrule
   {CUB} & 85.83$\pm$1.97 & 82.00$\pm$3.15 & 85.50$\pm$1.39 & 77.67$\pm$2.99 & 90.83$\pm$3.23 \\
   \midrule
   {PIE} & 80.88$\pm$0.95 & 80.59$\pm$1.52 & 82.35$\pm$1.04 & 76.42$\pm$1.37 & 91.91$\pm$0.11 \\
   \midrule
   {Caltech101} & 90.50$\pm$0.00 & 88.84$\pm$0.41 & 89.97$\pm$0.41 & 88.11$\pm$0.40 & 92.93$\pm$0.20 \\
   \midrule
   {Scene15} & 55.77$\pm$0.22 & 54.85$\pm$1.00 & 55.03$\pm$0.34 & 53.82$\pm$0.24 & 67.74$\pm$0.36 \\
   \midrule
   {HMDB} & 54.34$\pm$0.75 & 46.73$\pm$0.97 & 49.16$\pm$1.07 & 49.12$\pm$1.01 & 65.26$\pm$2.52 \\
   \bottomrule
   \end{tabular}%
   }
 \label{tab:cca}%
\end{table}%
{\subsection{Experimental analysis of removing view}}
\begin{table}[htbp]
\tiny
\centering
 {
    \caption{Experimental results of manually removing view.}
    \begin{tabular}{c|cccc|cccc}
    \toprule
    Data  & \multicolumn{4}{c|}{PIE} & \multicolumn{4}{c}{Scene15} \\
    \midrule
    Removed view &View 1     &View 2     &View 3     & None  &View 1     &View 2     &View 3     & None \\
    %ACC   & 87.06$\pm$0.36 & 89.56$\pm$0.55 & 84.12$\pm$0.36 & 91.99$\pm$1.01 & 63.65$\pm$0.43 & 55.66$\pm$0.55 & 64.41$\pm$0.53 & 67.74$\pm$0.36 \\
    ACC   & 87.06$\pm$0.36 & 89.56$\pm$0.55 & 84.12$\pm$0.36 & 91.99$\pm$1.01 & 63.65$\pm$0.43 & 55.66$\pm$0.55 & 64.41$\pm$0.53 & 67.74$\pm$0.36 \\
    \bottomrule
    \end{tabular}%
  \label{tab:removing}%
  }
\end{table}%
{We conduct experiments by removing one view manually on two datasets (i.e., PIE and Scene15) associated with three views. We run 30 times and report the mean and standard deviation in terms of classification accuracy. The experimental results are shown in Table~\ref{tab:removing}. It is observed that by using more information, the performance of our model tends to be improved.}
\subsection{Details of the implementation}
{For our trusted multi-view classification (TMC) algorithm, we employ the fully connected networks for all datasets, and $l_2$-norm regularization is used with the value of the trade-off parameter being 0.0001. The Adam \citep{kingma2014adam} optimizer is used to train the network. 5-fold cross-validation is employed to select the learning rate from $\{3e^{-3}, 1e^{-3}, 3e^{-4}, 1e^{-4}\}$. For all datasets, $20\%$ of samples are used as test
set. The hyperparameter $\lambda_t$ in
Eq.~\ref{eq:la} slowly increases from 0 to 1. For all methods, we run 30 times to report the mean values and standard deviations.}

\subsection{End-to-end experiment}
{Although the effectiveness has been well validated on datasets with multiple types of features, we further conduct experiments in an end-to-end manner on multi-modal datasets. The UMPC-FOOD101 dataset consists of 86,796 samples scraped from web pages, each sample is described with both image and text \citep{wang2015recipe}. Similarly to existing work \citep{kiela2019supervised}, there are 60,101 samples used as the training set, 5,000 samples used as the validation set, and the remaining 21,695 samples are used as the test set. The samples of NYUD (RGB-D) dataset are collected by cropping out tight bounding boxes around instances of 19 object classes in the NYUD \citep{silberman2012indoor} dataset. There are 4,587 samples, and each sample is composed of both RGB and depth modalities. We use 2,186, 1,200, and 1,201 samples as the training set, validation set, and test set respectively.}  {In this experiment, we used the original datasets which contain noise due to the collection process.}

\begin{figure}[!htbp]
\begin{floatrow}
\ffigbox{%
  \includegraphics[height=4.5cm]{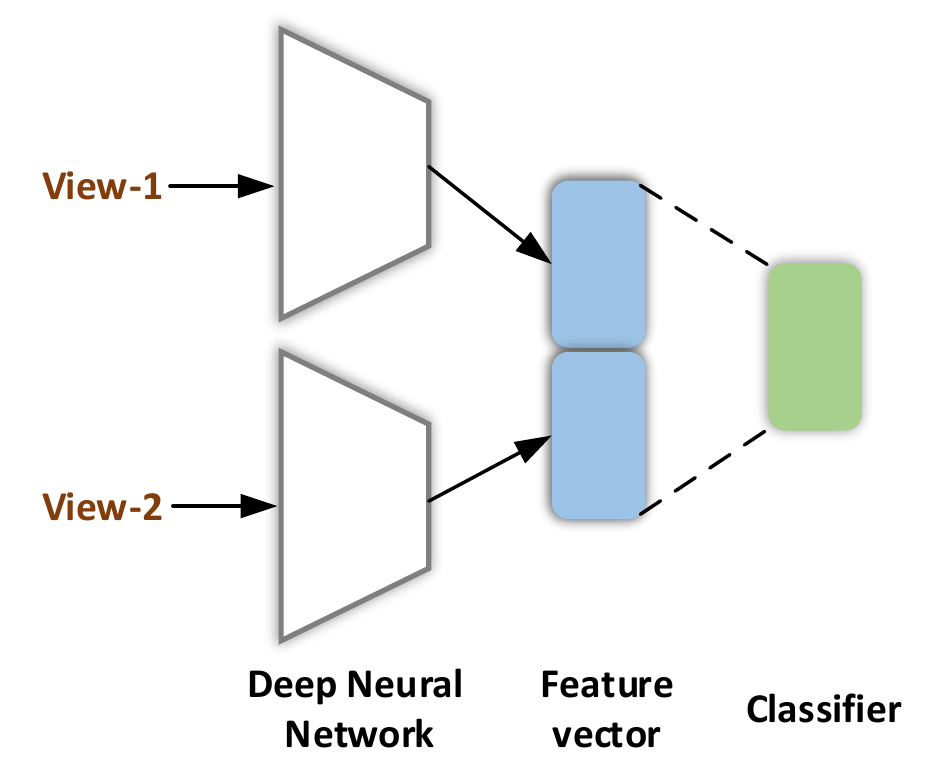}
}{%
  \caption{Feature fusion of two modalities in end-to-end manner.\label{fig:framework3}}%
}
\capbtabbox{%
 \footnotesize
    \begin{tabular}{ccc}
    \toprule\
    method & Food101 & NYUD \\
    \midrule
    \scriptsize{MCDO (best view)} & 74.87$\pm$1.04 & 55.83$\pm$1.67 \\
    \scriptsize{DE (best view)} & 83.27$\pm$0.36 & 57.95$\pm$1.42 \\
    \scriptsize{UA (best view)} & 84.49$\pm$0.34 & 56.04$\pm$1.16 \\
    \scriptsize{EDL (best view)} & 87.40$\pm$0.48 & 57.81$\pm$0.79 \\
    \midrule
    \scriptsize{MCDO (feature fusion)}  & 79.49$\pm$1.54 & 58.73$\pm$2.34 \\
    \scriptsize{DE (feature fusion)}    & 86.45$\pm$0.44 & 62.47$\pm$1.86 \\
    \scriptsize{UA (feature fusion)}    & 88.50$\pm$0.47 & 63.04$\pm$1.37 \\
    \scriptsize{EDL (feature fusion)}   & 90.50$\pm$0.32 & 64.49$\pm$1.14 \\
    \midrule
    Ours  & \textbf{91.30$\pm$0.21} & \textbf{66.24$\pm$0.69} \\
    \bottomrule
    \end{tabular}%
}{%
  \caption{Evaluation of the classification performance on Food101 and NYUD.
  \label{tab:2}}%
}
\end{floatrow}
\end{figure}

{For UMPC-FOOD101, we employ ResNet-152 \citep{he2016deep} pre-trained with ImageNet and BERT \citep{devlin2018bert} as the backbone network for image and text respectively. For NYUD (RGB-D), ResNet-50 \citep{he2016deep} pre-trained with ImageNet is used as the backbone network for depth and RGB images. For comparison algorithms, we report the results of the best performing views, and furthermore, we concatenate the outputs of the two backbone networks as the input of the classifier as shown in Fig.~\ref{fig:framework3}. For all algorithms, due to the randomness involved, we run 10 times to report the mean accuracy and standard deviation. The Adam \citep{kingma2014adam} optimizer is employed for all methods with learning rate decay.}

\textbf{Quantitative experimental results.} The experimental results on the above datasets are shown in Table~\ref{tab:2}. We report the accuracy of each compared algorithm with the best-performing view (`best view' in Table~\ref{tab:2}) and combined views (`feature fusion' in Table~\ref{tab:2}). According to Table~\ref{tab:2}, our algorithm outperforms other methods on all datasets. For example, on the Food101 dataset, our method achieved $91.3\%$ in terms of accuracy, compared with $90.5\%$ from the second performer. The performance of using the best single view is relatively low.

%-----------------------------

\begin{table*}[b]
    \centering
    \small
    \renewcommand\arraystretch{1.1}
    \begin{tabular}{p{0.5cm}p{7cm}p{3cm}p{1.5cm}}
        \toprule
        \textbf{Index}&\textbf{Text} & \textbf{Image} & \textbf{Label}\\
        \midrule
        1&
        \raisebox{-.9\height}{\includegraphics[width=210pt]{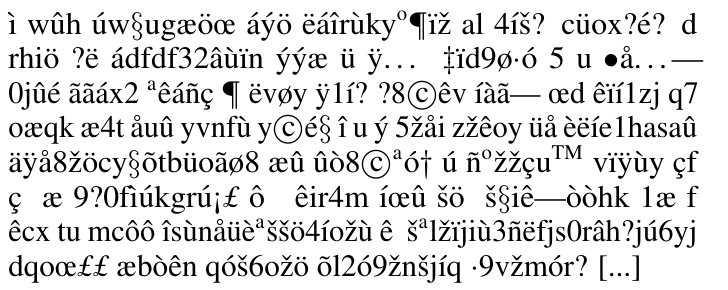}}
        %‹      ì½ wûh úw÷§€ugæöœ áýð«ö ëáîrùkyº¶ïž a¢l 4íš? ÷cüox?é? d r¤hiö ?»ë ádfdf¼32âùƒ×ïn þýýæ¨ þü ÿ… ‡ïd9ø·ó 5 u •å…| 0jûé ããáx2 ªêáñç¬ ¶ ´ëvøy ÿ1í? ?8©êv íàã| œd ðêïí1zj q7² þo„æq×k æ4t åuû ½yv¢nfù y©é§ î u ý 5žåi zžêoy üå ðèë¼íe1hasa¨û¼ ä‹ÿå8žöcy§õtbüoãø8 æû¼ û‹ò8©ªó† ú ¥ñºžžçu™ vïÿùy çf ‹ç ¼ æ 9?0fµìúkgr¢ú›¡»£¤ ô µ þ êir4m íœû šö ¤ š§²iê|òòhk ½1æ f êcx tu mcô÷ô îsù´²nåüèªššö4ío²žù ê ¦ šªlžïji‹ù3ñðëfjs0râ±h?jú6yj dqoœ££ æb‰ò±ê³¦n ¬qóš6ožö õl2ðó9žnšj¤íq ·²9vžmór? [...]
        & \raisebox{-.9\height}{\includegraphics[width=80pt]{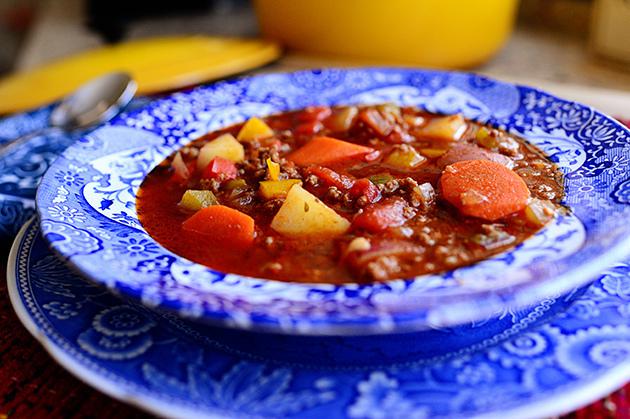}} & Hamburger\\\midrule
        2&log in sign up recipe box shopping list my profile my inbox account settings log out add a recipe looking for your public and private recipes or recipes with notes? they re now in your new recipe box learn more introducing the new food com app getting dinner on the table just got a whole lot easier 31 kitchen tips that ll change your life everything you need to conquer life in the kitchen don t miss secret ingredient burgers grilled chicken 15 ways and fun hot dog recipes looking for more? [...]& \raisebox{-.9\height}{\includegraphics[width=80pt]{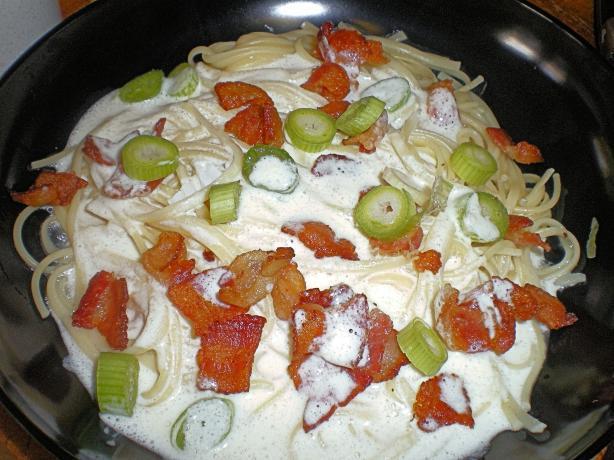}} & Spaghetti carbonara\\\midrule
        3&ack to channel options | sign out welcome sign in options | sign up account info make aol my homepage aol com aol com mail aol mail view all mail compose aol on favorites favorites search for videos playlists aol originals all originals so much more in short park bench hardwired cityballet candidly nicole the future starts here the restart project anthony eats america fatherhood acting disruptive the sartorialist flat out more shows [...] & \raisebox{-.9\height}{\includegraphics[width=80pt]{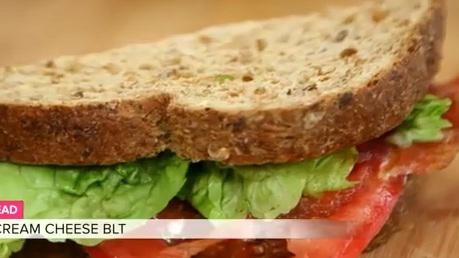}} & Grilled cheese sandwich\\\midrule
        4&shows featured shows barbecue addiction bobby s basics barefoot contessa diners drive ins and dives farmhouse rules giada at home restaurant impossible the kitchen the pioneer woman trisha s southern kitchen all shows a z videos watch food network star salvation full episodes watch food network star watch beat bobby flay watch full episodes chopped diners drive ins and dives the pioneer woman more full episodes more shows tv schedule chefs ree drummond summer party picks cooking for cowboys guy fieri super satisfying burgers dedicated fans of ddd alton brown giada de laurentiis bobby flay ina garten damaris phillips robert irvine more [...] &
        \raisebox{-.9\height}{\includegraphics[width=80pt]{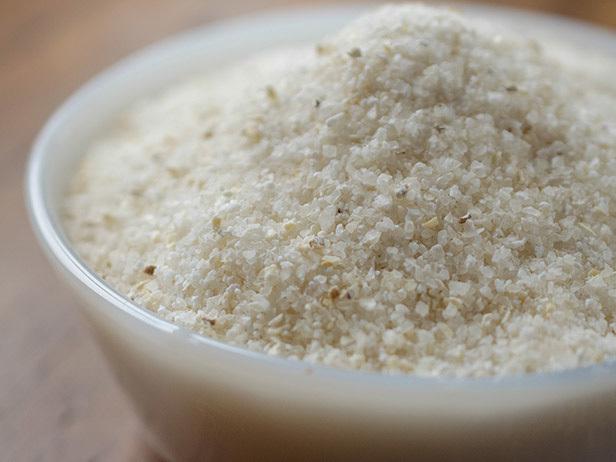}} & Shrimp and grits\\\midrule
        5&the heritage cook ® sharing traditions one recipe at a time by email home about contact disclaimers the heritage cook working together gluten free gf baking tips hints gf flour recipes gf resources websites how to build a gf flour blend from scratch heritage store recipe lists appetizers snacks … baked goods breads … food gifts breakfast brunch … beverages chocolate mondays … desserts main courses sauces marinades dry rubs … sides vegetables soups … salads dressings the bookcase baking shelf cuisine diet ingredient shelf general cooking and reference shelf the pantry chocolate faqs [...] &
        \raisebox{-.9\height}{\includegraphics[width=80pt]{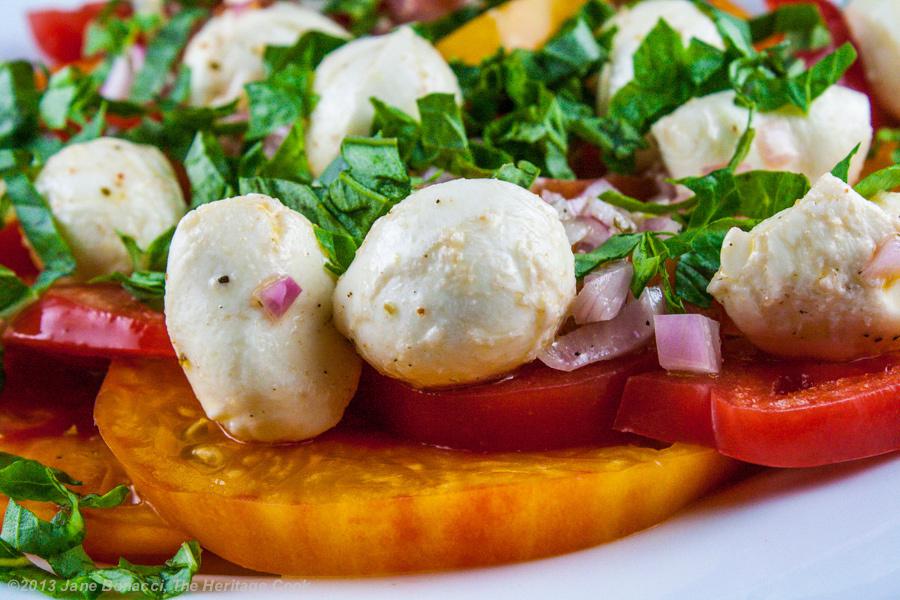}} & Caprese salad
     \\\bottomrule
    \end{tabular}
    \caption{Examples with the highest (top 5) prediction uncertainty on Food101. {These 5 examples are misclassified.}}
    \label{tab:examples1}
\end{table*}

\textbf{Qualitative experimental results.}
We also present typical examples (from UMPC-Food101) with the highest and lowest uncertainty in prediction in Table~\ref{tab:examples1} and Table~\ref{tab:examples2}, respectively. Since the UMPC-Food101 dataset is scraped from web pages, for a few samples the corresponding text descriptions are meaningless (\emph{e.g.}, the 1st one in Table~\ref{tab:examples1}). At the same time, some images are also quite challenging to classify (\emph{e.g.}, the 2nd and the 4th in Table~\ref{tab:examples1}). For these  samples that can be considered as out-of-distribution ones, it is difficult to obtain evidence related to classification. As expected, these samples are assigned with high uncertainty (top 5) by our algorithm. In contrast, Table~\ref{tab:examples2} presents the top 5 samples with lowest uncertainty. We can find that for these samples the labels are usually explicitly mentioned in the text and the images also clearly show the characteristics of `macarons'. %Moreover, the labels of the most confident samples are macarons, which shows that the model has higher confidence in categories that are easier to distinguish. Correspondingly, in Table~\ref{tab:examples1}, we can find that the most uncertain samples have different categories and are difficult to distinguish.
\begin{table*}[!htbp]
    \centering
    \small
    \renewcommand\arraystretch{1.1}
    \begin{tabular}{p{0.5cm}p{7cm}p{3cm}p{1.5cm}}
        \toprule
        \textbf{Index}&\textbf{Text} & \textbf{Image} & \textbf{Label}\\
        \midrule
        1&gwen s kitchen creations delicious baked goods created in the kitchen for oscar \textbf{macaron} tips and tricks and a recipe by gwen on | i ve made french \textbf{macarons} many many times my first few attempts ended in misery broken shells cracks hollows no feet everything that could go wrong in a bad \textbf{macaron} did however i never stopped trying who knows how much pounds of almond flour meal and powdered sugar i sieved but it was not in vain i learned so much from all these [...] & \raisebox{-.9\height}{\includegraphics[width=80pt]{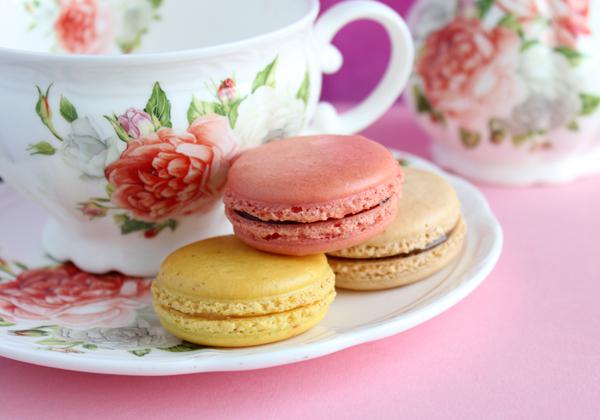}} & Macarons\\\midrule
        2&the pleasure monger telling it as it is menu skip to content home about me advertising collaborative opportunities using my content contact me tag archives \textbf{macaron} recipe sunflower seed \textbf{macarons} with black truffle salted white chocolate ganache 37 replies when the very talented and prolific shulie writer of food wanderings approached me on twitter to do a guest post for her tree nut free \textbf{macaron} series [...]& \raisebox{-.9\height}{\includegraphics[width=80pt]{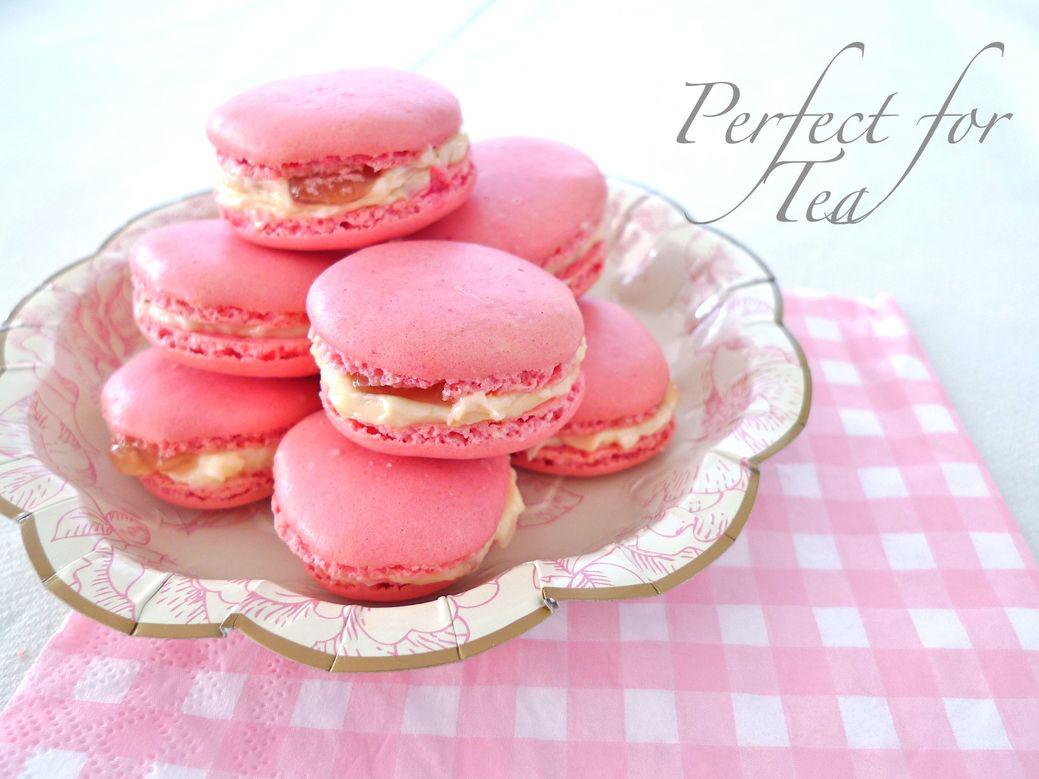}} & Macarons\\\midrule
        3&it s free and you can unsubscribe at any time submit your email address below and we ll send you a confirmation message right away approve with one click and you re done for more information click here get it delivered selected posts best new pastry chef why weight total eclipse of the tart homemade sprinkles about bravetart monday october 24 2011 \textbf{macarons} are for eating tweet when i've posted about \textbf{macarons} before in \textbf{macaron} mythbusters [...] & \raisebox{-.9\height}{\includegraphics[width=80pt]{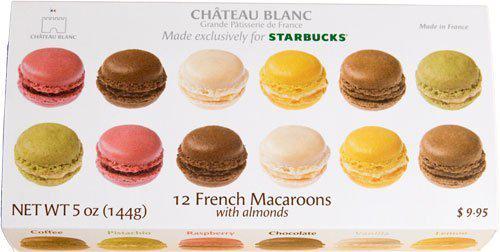}} & Macarons \\\midrule
        4&tastespotting features the delicious life facebook twitter tastespotting want to submit something delicious? login or register first browse by date | category | popularity | random tag \textbf{macarons} daydreamerdesserts blogspot com helping end childhood hunger one \textbf{macaron} at a time s mores banana cream pie apple pie oatmeal raisin piña colada \textbf{macarons} 83305 by tavqueenb save as favorite bakingupforlosttime blogspot com \textbf{macaron} party my first attempt at french \textbf{macarons} [...] &
        \raisebox{-.9\height}{\includegraphics[width=80pt]{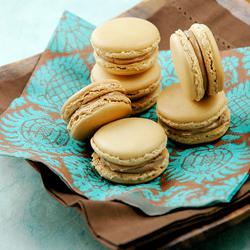}} & Macarons\\\midrule
        5&blue ribbons recipe contests giveaways meet the kitchen crew cookbooks custom cookbooks member cookbooks members choice cookbooks my cookbooks coupons shop knickknacks gift memberships members choice cookbooks my kitchen my profile recipe box cookbooks menu calendar grocery list conversations notifications hide ad trending recipes rice krispy treat \textbf{macarons} rice krispy treat \textbf{macarons} 1 photo pinched 3 times grocery list add this recipe to your grocery list print print this recipe and money saving coupons photo [...] &
        \raisebox{-.9\height}{\includegraphics[width=80pt]{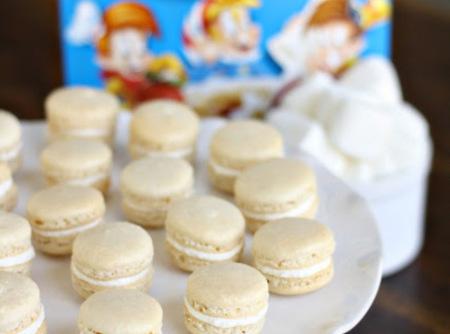}} & Macarons
     \\\bottomrule
    \end{tabular}
    \caption{Examples with the lowest (top 5) prediction uncertainty on Food101. {The above 5 samples are correctly classified.}}
    \label{tab:examples2}
\end{table*}

\end{document}